\documentclass[sn-basic,iicol]{sn-jnl}% Default with double column layout

\renewenvironment{table}[1][]%
{\tableorg[#1]%
\tablebodyfont%
\renewcommand\footnotetext[2][]{{\removelastskip\vskip3pt%
\let\tablebodyfont\tablefootnotefont%
\hskip0pt\if!##1!\else{\smash{$^{##1}$}}\fi##2\par}}%
}{\endtableorg}

\usepackage{graphicx}%
\usepackage{multirow}%
\usepackage{amsmath,amssymb,amsfonts}%
\usepackage{amsthm}%
\usepackage{mathrsfs}%
\usepackage[title]{appendix}%
\usepackage{xcolor}%
\usepackage{textcomp}%
\usepackage{manyfoot}%
\usepackage{booktabs}%
\usepackage[lined,boxed,commentsnumbered,ruled]{algorithm2e} %for algorithm
\usepackage{algorithmicx}%
\usepackage{algpseudocode}%
\usepackage{listings}%

\newtheorem{myDef}{Definition}

\begin{document}
	
\title[Article Title]{Automatic Construction of Parallel Algorithm Portfolios for Multi-objective Optimization}
	
	%%=============================================================%%
	%% Prefix	-> \pfx{Dr}
	%% GivenName	-> \fnm{Joergen W.}
	%% Particle	-> \spfx{van der} -> surname prefix
	%% FamilyName	-> \sur{Ploeg}
	%% Suffix	-> \sfx{IV}
	%% NatureName	-> \tanm{Poet Laureate} -> Title after name
	%% Degrees	-> \dgr{MSc, PhD}
	%% \author*[1,2]{\pfx{Dr} \fnm{Joergen W.} \spfx{van der} \sur{Ploeg} \sfx{IV} \tanm{Poet Laureate} 
		%%                 \dgr{MSc, PhD}}\email{iauthor@gmail.com}
	%%=============================================================%%
	
\author[1,2]{\fnm{Xiasheng} \sur{Ma}}
	
\author*[3]{\fnm{Shengcai} \sur{Liu}}\email{liusccc@gmail.com}

\author[4]{\fnm{Wenjing} \sur{Hong}}

\affil[1]{\orgdiv{School of Mechanical Engineering}, \orgname{ University of Science and Technology Beijing}, \orgaddress{ \city{Beijing}, \postcode{100083}, \country{China}}}
	
\affil[2]{\orgdiv{Beijing Key Laboratory of Research and Application for Robotic Intelligence of Hand-Eye-Brain Interaction}, \orgaddress{\city{Beijing}, \postcode{100083}, \country{China}}}

\affil[3]{\orgdiv{Centre for Frontier AI Research (CFAR)}, \orgname{Agency for Science, Technology and Research (A*STAR)}, \orgaddress{\country{Singapore}}}

\affil[4]{\orgdiv{Guangdong Key Laboratory of Brain-Inspired Intelligent Computation, Department of Computer Science and Engineering}, \orgname{Southern University of Science and Technology}, \orgaddress{\city{Shenzhen}, \postcode{518055}, \state{Guangdong}, \country{China}}}
	
	%%==================================%%
	%% sample for unstructured abstract %%
	%%==================================%%
	
\abstract{It has been widely observed that there  exists no universal best Multi-objective Evolutionary Algorithm (MOEA) dominating all other MOEAs on all possible Multi-objective Optimization Problems (MOPs).
In this work, we advocate using the Parallel Algorithm Portfolio (PAP), which runs multiple MOEAs independently in parallel and gets the best out of them, to combine the advantages of different MOEAs.
Since the manual construction of PAPs is non-trivial and tedious, we propose to automatically construct high-performance PAPs for solving MOPs.
Specifically, we first propose a variant of PAPs, namely MOEAs/PAP, which can better determine the output solution set for MOPs than conventional PAPs.
Then, we present an automatic construction approach for MOEAs/PAP with a novel performance metric for evaluating the performance of MOEAs across multiple MOPs.
Finally, we use the proposed approach to construct a MOEAs/PAP based on a training set of MOPs and an algorithm configuration space defined by several variants of NSGA-II.
Experimental results show that the automatically constructed MOEAs/PAP can even rival the state-of-the-art ensemble MOEAs designed by human experts, demonstrating the huge potential of automatic construction of PAPs in multi-objective optimization.}


%%\pacs[JEL Classification]{D8, H51}

%%\pacs[MSC Classification]{35A01, 65L10, 65L12, 65L20, 65L70}

\maketitle

\section{Introduction}
\label{Introduction}

Multi-objective Optimization Problems (MOPs) are a class of classical optimization problems, which widely exist in many real-world applications such as container terminal scheduling \citep{WOS:000574099800001}, vehicle routing \citep{LIU2021100927}, traffic engineering \citep{WOS:000582098300001}, and vehicle shape design \citep{WOS:000585987200008}.
% mobile edge computing \citep{LIN2022101163}, vehicle routing \citep{LIU2021100927}, image classification \citep{LIU2021100794}, image segmentation \citep{LIU2021100792}, portfolio investment \citep{ChenZD21}, and industrial design \citep{KumarWALMSD21}.
% expensive optimization problems \citep{num1}, robust optimization problems \citep{num2}, discrete optimization problems \citep{num3}, constrained optimization problems \citep{num4}, and large-scale continuous problems \citep{num5}.
Multi-objective Evolutionary Algorithms (MOEAs), as a kind of algorithms for solving MOPs, have attracted great attention over the past few decades and have shown good performance \citep{num26,ZhangL07,ZitzlerK04,zhou2011multiobjective,num5}.
Generally, MOEAs follow the generational framework where a population of solutions evolve from one generation to the next.
In the evolution process, evolutionary operators (e.g., mutation and crossover) are used to generate offsprings through the parents, while environmental selection operators are used to select elite individuals for the next-generation evolution.

Despite the tremendous success achieved by MOEAs, it has been widely observed \citep{num5, wu2019ensemble, num9, num7, num8,  num6} that there exists no universal best MOEA dominating all other MOEAs on all possible MOPs.
Instead, different MOEAs are good at solving different MOPs due to their abilities on exploitation and exploration \citep{WangYLZWCC19}.
For example, MOEAs equipped with polynomial-based mutation (PM) \citep{coello2007evolutionary} are good at searching in a local area, while those using simulated binary crossover (SBX) \citep{GohTLC10} and differential evolution (DE) \citep{num27,DasS11} are capable of exploring the search space globally.
Hence, in order to achieve a better overall performance for a diverse range of MOPs, it is natural and intuitive to combine the advantages of different MOEAs.
One notable series of research efforts following this idea are some ensemble strategies \citep{ num9, num7, num8,  num6, VNEF,  EFPD, TAOS, EMBBO, ENSMOEAD, HRISE, PAPHH}, which adaptively allocate computational resources to MOEAs equipped with different operators when solving a MOP.

Apart from them, and from a more general perspective of problem solving, there is an effective technique that exploits the complementarity between different algorithms by including them into a so-called algorithm portfolio (AP).
To utilize an AP to solve a problem, Tang~et al.~\citep{PengTCY10,TangPCY14} proposed a simple and popular strategy, called parallel algorithm portfolio (PAP), that runs all member algorithms in the portfolio independently in parallel to get multiple solutions.
Then, the best solution will be taken as the final output of the PAP. And Parsopoulos~et al.~\citep{parsopoulos2022parallel} conducted research on resource allocation based on adaptive pursuit strategy in PAP.

Although a PAP would consume more computational resources than a single algorithm, it has three important advantages. First, PAPs are easy-to-implement because they do not necessarily require any resource allocation since each member algorithm is simply assigned the same amount of resources.
	Second, the performance of a PAP on any problem is the best performance achieved among its member algorithms on the problem.
	In other words, a PAP could achieve much better overall performance than any of its member algorithms.
	Third, considering the tremendous growth of parallel computing architectures \citep{num15} (e.g., multi-core CPUs) over the last few decades, leveraging parallelism has become very important in designing effective solvers for hard optimization problems \citep{GebserKNS07a,RalphsSBK18,num13,num14,LiuTY22}.
	PAPs employ parallel solution strategies and thus allow using modern computing facilities in an extremely simple way.

It is conceivable that any PAP's effectiveness relies heavily on the diversity and complementarity among its member algorithms.
In other words, manual construction of high-quality PAPs is generally a challenging task, requiring domain experts (with a deep understanding of both algorithms and problems) to explore the vast design space of PAPs, which cannot be done manually with ease \citep{HamadiW13}.
As an alternative, Tang and Liu~\citep{num13,num14} proposed a general framework, called automatic construction of PAPs, that seeks to automatically build PAPs by selecting the member algorithms from an algorithm configuration space, with the goal of optimizing the performance of the resulting PAP on a given problem set (called training set).
%In ref~\citep{ACsurvey}, some types of search strategies are displayed for the configuration search, e.g., model-free methods, model-based methods, theoretical guarantees, and instance-specific methods. Among them, the model-based method, SMAC 3 \citep{lindauer2022smac3}, was adopted by Tang and Liu~\citep{num13,num14}.
The framework have been shown effective in building high-performance PAPs for a variety of problems such as the Boolean Satisfiability Problem (SAT) \citep{num13}, the Traveling Salesman Problem (TSP) \citep{LiuTY22,LiuYT2022}, the Vehicle Routing Problem (VRP) \citep{num14}, and Adversarial Attacks \citep{abs-2211-12713}.

However, to the best of our knowledge, the potential of automatic construction of PAPs has not been investigated in the area of multi-objective optimization.
Considering its excellent performance on the above-mentioned problems and the practical significance of MOPs, studying how to utilize it to solve MOPs is thus valuable.
In this work, we focus on automatically building PAPs for continuous MOPs.
On the other hand, as a general framework, appropriately instantiating automatic PAP construction for a specific problem domain is non-trivial.
Specifically, it requires careful designs of the algorithm configuration space and the performance metrics used in the construction process \citep{num14}.

The main contributions of this work can be summarized as follows.
\begin{enumerate}
	\item Taking the characteristics of MOPs into account, we propose a novel variant form of PAP for MOPs, dubbed MOEAs/PAP. Its main difference from conventional PAPs lies in the way of determining the final output. MOEAs/PAP would compare the solution sets found by member algorithms and the solution set generated based on all the solutions found by member algorithms, and finally output the best solution set.
	\item We present an automatic construction approach for MOEAs/PAP with a novel metric that evaluates the performance of MOEAs/PAPs across multiple MOPs.
	\item Based on a training set of MOPs and an algorithm configuration space defined by several MOEAs, we use the proposed approach to construct a MOEAs/PAP.
	Experimental results show that MOEAs/PAP significantly outperforms the foundation algorithms and can even rival the state-of-the-art ensemble MOEAs designed by human experts.
	Such promising results indicate the huge potential of automatic construction of PAPs in multi-objective optimization.
\end{enumerate}

The remainder of this paper is organized as follows.
Section \ref{Preliminaries and Related work} presents the preliminaries and briefly reviews the related works.
Section \ref{MOEAs/PAP} gives the variant form of PAP for MOPs.
The automatic construction approach, as well as the algorithm configuration space, the training set and the performance metric used in construction, are presented in Section \ref{automatic construction}.
Section \ref{experiments} presents the experimental study.
Finally, Section \ref{conclusion} concludes the paper.

\section{Preliminaries and Related Work}
\label{Preliminaries and Related work}

\subsection{Multi-objective Optimization Problems}

Without loss of generality, in this work we assume optimization problems take the minimization form.
MOPs are optimization problems with multiple objectives, defined as follows:
\begin{equation}
	\label{MOPs}
	\begin{aligned}
		min \ F(\textit{\textbf{x}}) &= [f_{1}(\textit{\textbf{x}}),f_{2}(\textit{\textbf{x}}), \ldots,f_{m}(\textit{\textbf{x}})] \in \Pi \\
		\textit{\textbf{x}} &= [\textit{x}_{1}, \textit{x}_{2}, \ldots, \textit{x}_{n}] \in D
	\end{aligned},
\end{equation}
where $\textbf{x}$ is the decision vector which consists of $n$ decision variables.
The objective vector $F: D \rightarrow \Pi $ consists of $m$ objectives; $D \subseteq \textit{R}^{n} $ and  $\Pi \subseteq \textit{R}^{m} $ denote the decision space and the objective space, respectively.
The objectives in Eq.~\eqref{MOPs} are often in conflict with each other. 
That is, the improvement of one objective may lead to the deterioration of another.
This gives rise to a set of optimal solutions (largely known as Pareto-optimal solutions), instead of a single optimal solution.
The concept of Pareto-optimal is defined as follows.

\begin{myDef}
	\label{domination}
	
	Given two decision vectors $\textit{\textbf{u}}, \textit{\textbf{v}}$ and their corresponding objective vectors $F(\textit{\textbf{u}}), F(\textit{\textbf{v}})$. $\textit{\textbf{u}}$ dominates $\textit{\textbf{v}}$ (denoted as  $\textit{\textbf{u}} \prec \textit{\textbf{v}}$), if and only if $ \forall i \in \{1, \ldots, m\} ,  f_{i}(\textit{\textbf{u}}) \le f_{i}(\textit{\textbf{v}})$ and $ \exists i \in \{1, \ldots, m\} ,  f_{i}(\textit{\textbf{u}}) < f_{i}(\textit{\textbf{v}})$.
	
\end{myDef}

\begin{myDef}
	\label{domination_set}
	A solution ${\textbf{x}}^{*}$ is Pareto-optimal if and only if there exixts no $\textbf{u} \in \Omega$ such that $\textbf{u} \prec {\textbf{x}}^{*}$. The set of all Pareto-optimal solutions is called the Pareto set. The corresponding objective vector set of the Pareto set is called the Pareto front.
\end{myDef}

There exist a large body of benchmark sets for MOPs.
%To facilitate the study of the MOPs, researchers have proposed many benchmark problems. 
Among them, Zizler-Deb-Thiele (ZDT) \citep{num16}, Deb-Thiele-Laumanns-Zizler (DTLZ) \citep{num17}, walking-fish-group (WFG) \citep{num18}, CEC 2009 competitions (UF) \citep{num19} and many-objective optimization problems (MaOP) \citep{MaOP} are commonly used in the literature.
These problems are characterized by multimodality, co-replicated Pareto sets and Pareto fronts, separability between decision variables, and correlation between decision variables and objectives, deception, and epistasis \citep{num18, num20}. These are of varying difficulty and complexity and cover a variety of real-world problems.
Moreover, the commonly used metrics for evaluating the solutions on MOPs include Inverted Generational Distance (IGD)\citep{num21} and Hypervolume (HV)\citep{num22, num23}.

\subsection{Multi-objective Evolutionary Algorithms}

There are a large body of MOEAs, e.g. NSGA-II \citep{num26}, MOEA/D \citep{MOEAD}, SPEA2 \citep{SPEA2} and IBEA \citep{IBEA}. 
They represent different selection methods, which can be combined with different generation operators to generate a variety of algorithms. Genetic Algorithms (GAs) \citep{num26,ZhangL07} and Differential Evolution (DE) Algorithms \citep{num25} are two mainstream types of generation operators.
%GA and DE, as classical algorithms, are commonly used. So in this paper, we use them for research.
GA is based on biological inspiration and generates new offspring by imitating the crossover and mutation process in biological genetic evolution.
One of the most classic and widely used variation strategies is composed of a simulated binary crossover (SBX) operator and a polynomial mutation (PM) operator \citep{num24}.
Specifically, supposing the two parent individuals are $\textit{\textbf{x}}^1=(x^1_1,x^1_2,\ldots,x^1_n)$ and $\textit{\textbf{x}}^2=(x^2_1,x^2_2,\ldots,x^2_n)$, and the generated offspring are $\textit{\textbf{c}}^1=(c^1_1,c^1_2,\ldots,c^1_n)$ and $\textit{\textbf{c}}^2=(c^2_1,c^2_2,\ldots,c^2_n)$.
SBX and PM operate as follows.

\textit{\textbf{SBX:}}
\begin{equation}
	\begin{aligned}
		c^1_i &= 0.5\times [(1+\beta)\cdot x^1_i + (1-\beta)\cdot x^2_i ]\\
		c^2_i &= 0.5\times [(1-\beta)\cdot x^1_i + (1+\beta)\cdot x^2_i ]
	\end{aligned},
	\label{SBX1}
\end{equation}
where $\beta$ is a parameter, defined as follows:
\begin{equation}
	\beta = 
	\begin{cases}
		(r\times 2)^{1/(1+\eta)} &\text{if} \ r \leq 0.5\\
		(1/(2-r\times 2))^{1/(1+\eta)} &\text{otherwise}
	\end{cases},
	\label{SBX2}
\end{equation}
where $r$ is a random value within $[0, 1]$ and $\eta$ is a parameter representing the similarity between the offspring individual and the parent individual. The larger the value, the higher the similarity.

\textit{\textbf{PM:}}
\begin{equation}
	\begin{aligned}
		c^1_i = x^1_i+\Delta \cdot (u_i-l_i)
	\end{aligned},
	\label{PM1}
\end{equation}
$\Delta $ is defined as follows:
\begin{equation}
	\Delta = 
	\begin{cases}
		(2r+(1-2r)\delta_1^{\eta +1})^{1/(\eta +1)-1}-1 &\text{if}\ r\leq 0.5\\
		((4r-1)+(2r-1)\delta_2^{\eta +1})^{1/(\eta +1)} &\text{otherwise}
	\end{cases},
	\label{PM2}
\end{equation}
where $\delta_1 =( u_i - x^1_i)/(u_i-l_i) $ and $\delta_2 = ( x^1_i - l_i)/(u_i-l_i)$, $r$ is a random parameter within [0, 1], $u_i$ and $l_i$ are the upper and lower bounds of the $i$-th dimension variable, respectively. $\eta$ is a parameter that representing the similarity between the offspring individual and the parent individual.
The larger the value, the higher the similarity.

Different from GAs, DEs mainly perform gradient estimation through differential mutation operators to generate offspring. Over the past few decades, various types of differential mutation operators have been proposed.
\citet{num25} summarized them in Eq.~\eqref{rand p}--Eq.~\eqref{current ot best p}:

\begin{figure*}
	\textit{\textbf{rand/p:}}
	\begin{equation}
		c^i_j = 
		\begin{cases}
			x^{r_3}_j+F\cdot \sum_{l=1}^{p}(x^{r_1,l}_j-x^{r_2,l}_j)  &\text{if}\ r \leq CR\ \text{or}\ j=j_r \\
			x^i_j, & \text{otherwise}
		\end{cases},
		\label{rand p}
	\end{equation}
	
	\textit{\textbf{best/p:}}
	\begin{equation}
		c^i_j = 
		\begin{cases}
			x^{best}_j+F\cdot \sum_{l=1}^{p}(x^{r_1,l}_j-x^{r_2,l}_j),  &\text{if} \ r\leq CR\ \text{or}\ j=j_r \\
			x^i_j, & \text{otherwise}
		\end{cases},
		\label{best p}
	\end{equation}
	
	\textit{\textbf{current-to-rand/p:}}	
	\begin{equation}
		\small
		c^i_j =
		\begin{cases}
			x^i_j + K \cdot (x^{r_3}_j - x^i_j) + F\cdot \sum_{l=1}^{p}(x^{r_1,l}_j-x^{r_2,l}_j) & \text{if}\ r\leq CR\ \text{or}\ j=j_r\\
			x^i_j & \text{otherwise}
		\end{cases},
		\label{current to rand p}
	\end{equation}
	
	\textit{\textbf{current-to-best/p:}}
	\begin{equation}
		\small
		c^i_j =
		\begin{cases}
			x^i_j + K \cdot (x^{best}_j - x^i_j) + F\cdot \sum_{l=1}^{p}(x^{r_1,l}_j-x^{r_2,l}_j) ,&\text{if}\ r\leq CR\ or\ j=j_r \\
			x^i_j ,& \text{otherwise}
		\end{cases}.
		\label{current ot best p}
	\end{equation}
\end{figure*}

In Eqs.~\eqref{rand p}--\eqref{current ot best p}, $p$ is the number of pairs of solutions used to compute the differences in the mutation operator, $c^i_j$ is the $j$-th dimension variable of the $i$-th offspring, $x^{r_3}_j$ is the $j$-th dimension variable of the donor solution chosen at random, $x^{best}_j$ is the $j$-th dimension variable of the best solution in the population as the donor solution, $x^i_j$ is the $j$-th dimension variable of the current parent, $x^{r_1,l}_j$ and $x^{r_2,l}_j$ are the $j$-th dimension variable of the $p$-th pair to compute the mutation differential, $F$, $K$, and $CR$ are parameters in the DE algorithm.

In additional, Coello et al. \citep{MOPSO} proposed a Multi-Objective Particle Swarm Optimization algorithm (MOPSO) for MOPs based on the particle swarm optimization algorithm. Later generations improved MOPSO, and proposed OMOPSO \citep{OMOPSO} and SMPSO \citep{SMPSO} respectively.

The main contribution of OMOPSO is to introduce a mutation mechanism on the basis of MOPSO. By dividing members into three parts and adopting different mutation strategies (uniform mutation, non-uniform mutation and no mutation), the search ability of the algorithm is enhanced. In addition to the mutation mechanism, SMPSO adopts the polynomial mutation algorithm on some particles, and also proposes a contraction coefficient to constrain the speed of the example.

\subsection{Ensemble Strategies}

In Ref.~\citep{wu2019ensemble}, there are a number of ensemble strategies for MOPs. Some are based on the adaptive adjustment strategy, e.g. ENS-MOEA/D~\citep{ENSMOEAD} and LDE~\citep{num9}. 
	The former, according to the artificially designed strategy, the performance of the algorithm is enhanced by adjusting a parameter, neighborsize, in the MOEAD algorithm. 
	While LDE regarded the parameter adaptation process as a Markov Decision Process (MDP). Through reinforcement learning, different parameter controllers are learned for different problems, such that they have different adaptive parameter selection methods for these problems.

Some of ensemble strategies contains multiple operators. These are cooperative (e.g. MSSEA~\citep{num7}, TOAs~\citep{num8} and EMBBO~\citep{EMBBO}) or competitive (e.g. MOEA/D-DRA~\citep{DRA}, MOEAs/MOE~\citep{num6} and VMEF~\citep{VNEF}).
	MSSEA takes the solutions generated by multiple operators as a whole and determined the local search strategy, global search direction and final convergence direction by analyzing the features in the entire objective space. Then different operators are used at different stages.
	TOAs constructs a solver that includes two complementary algorithms. The inferior algorithm optimizes itself by learning the strategy of the superior algorithm and gives feedback to fine-tune the superior algorithm. Finally, the desired result is output from the high-quality algorithm.
	EMBBO contains multiple similar member algorithms, and each algorithm independently generates offspring, and according to their common selection operator, the next generation that meets the corresponding number is obtained. Computing resources are evenly allocated among member algorithms.
	For MOEA/D-DRA and MOEAs/MOE, both of them allocate the resources occupied by each operator in the next generation according to the proportion of offspring saved by each operator in the previous generation. MOEA/D-DRA only includes two operators, and applies only to MOEAD. While MOEAs/MOE can includes more operators, motivated by the well-known ensemble learning approach AdaBoost \citep{FreundS97}, and is not limited to the types of MOEAs (only one can be used at the same time).
	VMEF adopts a voting mechanism. All operators vote to select new offspring, and then update the number of votes held by each operator in the next generation according to the hit rate of operator voting.

Besides, there are a lot of ensemble methods based on hyper-heuristics, such as HRISE~\citep{HRISE} and PAPHH~\citep{PAPHH}.
	The hyper-heuristics automate the heuristic design process based on the structure of the problem to be solved \citep{wu2019ensemble}. 
	In these methods, MOEAs are considered as low-level heuristics. They design selection method based on different strategies to realize the selection of appropriate low-level heuristic algorithms for different problems.
	HRISE combines reinforcement learning, (meta) heuristic selection, and group decision-making as acceptance methods. While PAPHH is porposed based on multiple diversity mechanisms, and develops the perturbation adaptive pursuit strategy to improve the decision-making process.

Combining some ideas from existing work, Wang et al. \citep{EFPD} proposed a new algorithm (EF-PD) for the coexistence of cooperative and competitive strategies. Among it, multiple selection operators jointly selecte the population, and multiple generation operators compete with each other to allocate computing resources, so that the advantages of each part can be fully utilized to obtain optimal performance. While Dong et al. \citep{TAOS} believed that in these methods, the evaluation of operators was unfair, so they proposed the MOEA/D-TAOS algorithm. By segmenting the evolution process, each operator was independently evaluated in the test phase, and in the application phase, the current optimal operator is selected for calculation.

\subsection{PAPs and Automatic Construction of PAPs}

Essentially, a PAP \citep{LiuTY22,LiuYT2022} is a set of member algorithms that run in parallel:
\begin{equation}
	\begin{aligned}
		P=\left\{\theta_1,\theta_2,\ldots,\theta_k\right\}
	\end{aligned},
	\label{PAP}
\end{equation}
where $P$ is the PAP solver, $\theta_{i}$ is the $i$-th member algorithm of $P$, and $k$ is the number of member algorithms. 

Let $M(\theta, z)$ denote the performance of an algorithm $\theta$ on a problem $z$ (e.g., a MOP) in terms of the performance metric $M$ (e.g., HV for MOPs).
Then, the performance of $P$ on a problem $z$, denoted as $\Omega(P, z)$, is the best performance achieved among the member algorithms of $P$ on $z$:
\begin{equation}
	\Omega(P,z)=\max_{\theta \in P}M(\theta,z).
	\label{performance}
\end{equation}
Without loss of generality, we assume for $M$, the larger the better.
Let $Z$ denote a set of problems.
The performance of $P$ on $Z$, denoted as $\Omega(P, Z)$, is an aggregated value of the performance of $P$ on all the problems in $Z$:
\begin{equation}
	\begin{aligned}
		\Omega(P,Z)=\frac{1}{|Z|} \cdot \sum_{z \in Z}\Omega(P,z).
	\end{aligned}
	\label{metric}
\end{equation}

The conventional way to construct $P$ is to manually choose member algorithms for it, as shown in \citep{PengTCY10,TangPCY14}, which is however non-trivial and tedious.
To address this issue, Tang and Liu~\citep{num13,num14} proposed the framework of automatic construction of PAPs.
Specifically, given a performance metric $M$, a training set $Z$, and an algorithm configuration space $\Theta$, the goal is to find at most $k$ algorithms from $\Theta$ to form a PAP $P^*$, which has the best performance on $Z$ in terms of $M$:
\begin{equation}
	\begin{aligned}
		P^*=\mathbb{\arg \max}_{P \subseteq \Theta \& |P|\leq k}\Omega(P,Z),
	\end{aligned}
	\label{automatic construction eq}
\end{equation}
where $\Omega(P,Z)$ is defined in Eq.~\eqref{metric}.
The results in \citep{num13,num14,LiuTY22,LiuYT2022} have shown that the PAPs automatically built by solving the problem in Eq.~\eqref{automatic construction eq} could achieve excellent performance on problems including SAT, TSPs and VRPs, which are even better than the algorithms designed by human experts.

\section{MOEAs/PAP for MOPs}
\label{MOEAs/PAP}
As stated in Eq.~\eqref{performance}, a PAP would return the best solution among the ones found by its member algorithms.
However, this could be problematic in multi-objective optimization since an MOEA outputs a set of, instead of a single one, Pareto-optimal solutions for a given MOP.
One way to mitigate this issue is to compare the solution sets found by the member algorithms in terms of the performance metric (e.g., HV) and finally returns the best solution set.

\begin{figure}[tbp]
	\centering
	\includegraphics[width=\columnwidth]{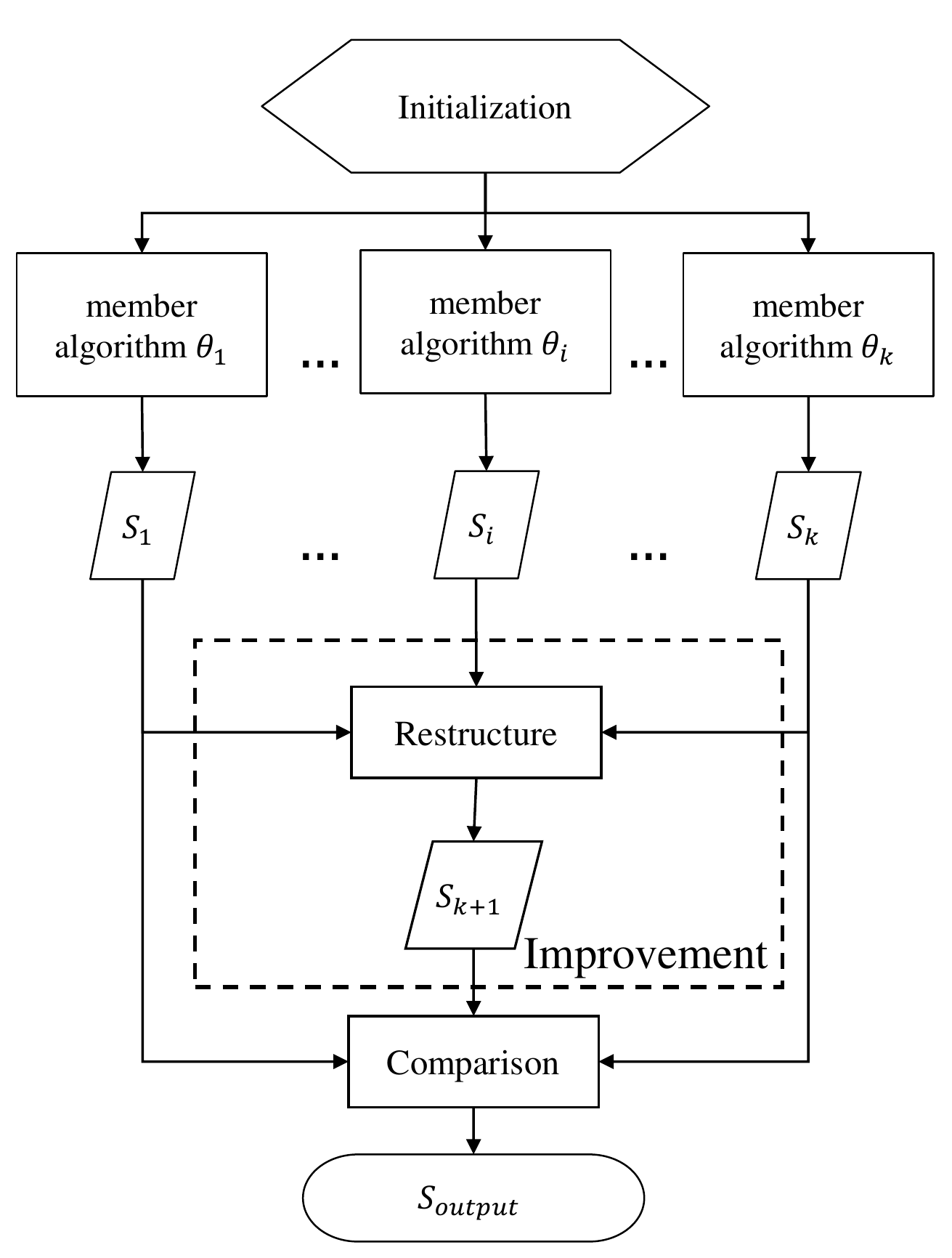}
	\caption{Illustrations of MOEAs/PAP, the variant form of PAP for MOPs. $S_i$ represents the solution set found by member algorithm $\theta_i$, and $S_{output}$ represents the solution set finally returned by MOEAs/PAP. The main difference between MOEAs/PAP and conventional PAPs lies in the way of determining the final output, as indicated by the dash box.}
	\label{framework}
\end{figure}

On the other hand, it is likely that the solution sets found by member algorithms contain solutions that do not dominate each other.
Thus, an alternative strategy is to first combine together the solutions contained in the sets generated by member algorithms, and then reconstruct a new Pareto-optimal solution set from them.
This procedure is called \textbf{Restructure}, where the quick-sorting algorithm of NSGA-II \citep{num26} is used to construct the new Pareto-optimal solution set.
Finally, all the solution sets found by member algorithms, as well as the one generated by \textbf{Restructure}, are compared, and the best set is returned as the output of MOEAs/PAP, as illustrated in Figure~\ref{framework}.

Formally, for a MOEAs/PAP, denoted as $P$, its performance on a MOP $z$ can be described as follows:
\begin{equation}
	\Omega(P,z)=\max \{ \max_{\theta \in P}M(\theta,z), M(\bar{\theta}, z)\},
	\label{new_performance}
\end{equation}
where $\bar{\theta}$ represents the \textbf{Restructure} procedure.
Note Eq.~\eqref{new_performance} is slightly different from Eq.~\eqref{performance} (i.e., the performance of conventional PAPs) due to the \textbf{Restructure} procedure.

\section{Automatic Construction of MOEAs/PAP}
\label{automatic construction}
As aforementioned, the member algorithms of a PAP could be automatically determined by solving the problem defined in Eq.~\eqref{automatic construction eq}.
In below we first introduce the algorithm configuration space $\Theta$, the training set $Z$, and the performance metric $M$; then we present the automatic construction approach for MOEAs/PAP.

\subsection{Algorithm Configuration Space $\Theta$ and Training set $Z$}
\label{sec:acs_trainingset}
%An algorithm configuration of a parameterized algorithm is a complete setting of the parameters of the algorithm, such that the algorithm?s behavior is completely speci?ed (up to randomization of the algorithm itself).
The algorithm configuration space $\Theta$ is defined by a number of parameterized algorithms $B_1, B_2, ..., B_c$ (called foundation algorithms).
Each foundation algorithm $B_i$ has a number of parameters, whose values control the behavior of $B_i$ and thus can largely affect its performance.
Therefore, $B_i$ taking different parameter values can actually be considered as different algorithms.
Let $\Theta_i$ denote the set of all unique algorithms obtained by taking all possible values of the parameters of $B_i$.
Then, the algorithm configuration space $\Theta$ is the union of $\Theta_1, \Theta_2, ... , \Theta_c$, i.e., $\Theta=\Theta_1 \cup  \Theta_2 \cup...\cup \Theta_c$.
In this work, the foundation algorithms are parameterized MOEAs, e.g., NSGA-II and MOEA/D.
The training set $Z$ is a set of MOPs, which is representative of the target problems to which the constructed MOEAs/PAP is expected to be applied.
The details of the algorithm configuration space $\Theta$ and the training set $Z$ are given in the experiments (see Sections~\ref{exp_instance_sets}-\ref{exp_automatic_construction}).

\subsection{Performance Metric $M$}
\label{sec:metric}

As stated in Eq.~\eqref{metric}, during the construction process, the performance of the PAP on the training set $Z$ is an aggregated value of its performance on all the problems in $Z$.
However, for MOPs, the commonly used IGD and HV cannot be directly used as the performance metric $M$ here.
The main reason is that for these two metrics, the performance of an MOEA on different MOPs cannot be directly aggregated (or compared) due to the different scales.

\begin{algorithm}[tbp]
	\LinesNumbered
	\SetKwInOut{Input}{input}
	\SetKwInOut{Output}{output}
	\Input{Algorithm configuration space $\Theta = \Theta_1 \cup  \Theta_2 \cup...\cup \Theta_c$, training set $Z$, maximum number of member algorithms $k$}
	\Output{the final PAP $P$}
	\tcc{-------Initialization-------}
	$ P \leftarrow \varnothing$;\\
	
	\While{$|P|<k$}
	{
		\tcc{-------Search in $\Theta$------}
		\For{$i = 1$ to $10$}
		{			
			$re = i\%c$;\\
			
			$\theta_i \leftarrow$ Find $\theta_i$ from $\Theta_{re}$ that maximizes $\frac{1}{|Z|} \cdot \sum_{z \in Z}(\Omega(P \cup \theta_i, z) - \Omega(P, z))$;\\
		}
		
		$\theta \leftarrow$ Find $\theta$ from $\{\theta_i\}$ that maximizes $\frac{1}{|Z|} \cdot \sum_{z \in Z}(\Omega(P \cup \theta, z) - \Omega(P, z))$;\\
		
		\leIf{$\Omega(P \cup \theta, Z) \leq \Omega(P, Z)$}{break}{$P \leftarrow P \cup \{\theta\}$}
		\tcc{----Simplification of $P$---}
		\While{true}
		{
			$P_{tmp} \leftarrow P$;\\
			\For{$\theta \in P$}
			{
				\lIf{$\Omega(P\setminus \{\theta\}, Z) \geq \Omega(P, Z)$}
				{
					$P \leftarrow P \setminus \{\theta\}$
				}
			}
			\lIf{$P_{tmp} = P$}{break}
		}
	}
	\Return{$P$}
	\caption{Automatic Construction of MOEAs/PAP}
	\label{automatic construction algorithm}
\end{algorithm}

Hence, we propose to re-scale HV values within $[0, 1]$, by using the following HV Ratio (HVR):
\begin{equation}
	hvr = \frac{hv}{hv^*},
	\label{HVR}
\end{equation}
%They cannot meet this requirement, so we need to construct a new performance metric to apply to the algorithm. The smaller value of IGD, the better, and tends to be zero. The larger value of HV, the better, and tends to be a fixed value related to the problem. HV with a fixed value related to the problem is more advantageous than IGD. We make improvements based on the HV and use the ratio to remove the difference between the problems. The new metric is named hypervolume percentage (HVR). The expression is as follows:
%\begin{equation}
%	hvr = \frac{hv}{hv^*},
%	\label{HVR}
%\end{equation}
where $hv$ is the HV value of the solution set found by an MOEA on a MOP, and $hv^{*}$ is the HV value of the optimal solution set of the problem.
Both $hv$ and $hv^{*}$ are calculated based on the same reference point in the objective space, which is obtained by taking the maximum on each objective achieved among the solution set found by the MOEA.
% Note here it is inappropriate to use the reference point of the optimal 
% normalized by the value of the current solution boundary.
% Compared with normalization using the value of the optimal boundary, it is more effective to evaluate the performance of all solutions.
However, there is still an issue in the actual use of HVR.
During the construction process of MOEAs/PAP, there is a large body of MOEAs which would be evaluated.
Since the reference point is up to the solution set found by the evaluated MOEA, given a MOP, $hv^{*}$ needs to be recalculated for every evaluated MOEA.
Considering the size of the optimal solution set is usually very large (typically larger than 1000), thus in the construction process it requires a large amount of calculations for $hv^{*}$ .

To address the above issue, we propose to use a fixed reference point for a given MOP, such that $hv^{*}$ needs to be calculated for the problem for only once.
One way is to use the the reference point taking the upper bounds (note that we assume MOPs take the minimization form) of the objective values.
However, this will cause the gap between $hv$ and $hv^*$ to be very small, in which case $hvr$ tends to be 1 and loses its evaluation ability.
Therefore, we calculate the HVs of the undominated space for the found solution set and the optimal solution set, and finally use the ratio of them to assess the performance of an MOEA on a MOP.

The new performance metric is named Inverted Hypervolume Ratio (IHVR):
\begin{equation}
	ihvr = \frac{hv_{all}-hv^*}{hv_{all}-hv},
\end{equation}
where $hv$ and $hv^*$ are calculated based on the reference point taking the upper bounds of the objective values, and $hv_{all}$  is the HV of the cube formed by the ranges of objective values of the MOP.
Thus, $hv_{all} - hv$ and $hv_{all} - hv^*$ are the HVs of the undominated space for the solution set found by the evaluated MOEA and the optimal solution set, respectively.
Note that $ihvr \in (0, 1]$, and a larger value of $ihvr$ is better.
IHVR can distinguish well between MOEAs with different performance, and can also save a lot of calculations.
In this work, IHVR is used as the performance metric $M$.

\subsection{Automatic Construction Approach}

Algorithm~\ref{automatic construction algorithm} presents the automatic construction approach for MOEAs/PAP.
Starting from an empty set (line 2), the approach constructs the PAP (denoted as $P$) iteratively.
Specifically, each iteration of the approach (lines 3-18) consists of two subsequent phases.
In the first phase, an existing automatic algorithm configuration tool, namely SMAC 3 \citep{lindauer2022smac3}, is used to search multiple solutions $\theta_i$ in different subspace $\Theta_{re}$ to
	find the algorithm that can improve the performance of the current PAP to the largest extent (line 7). These solutions are evenly distributed to the respective subspaces, and then the best one is inserted into $P$ (line 9).
This phase is similar to the commonly-used greedy approach in automatic construction of PAPs \citep{LiuTY22,LiuYT2022}.
Besides, we introduce a new phase, namely simplification, as the second phase in Algorithm~\ref{automatic construction algorithm}.
In this phase (lines 12-17), $P$ would be simplified by removing the member algorithms that do not contribute at all to its performance (meaning removing these algorithms has no effect on the performance of $P$ on the training set).
Considering the size of $P$ is bounded (line 3), removing the redundant  algorithms from $P$ is meaningful because this will leave space for new member algorithms that can improve the performance of $P$.
Finally, the approach would be terminated if the maximum number of member algorithms is reached (line 3), or the performance of $P$ does not improve with the inclusion of the found algorithm (line 10).

\begin{table}[tbp]
	%	\centering
	\caption{The problem benchmark sets used in the experiments.}
	\label{problem parameter}
	\scalebox{1.0}{
		\begin{tabular}{@{}lll@{}}
			\toprule
			\multirow{2}{*}{Problem} & Decision Vector & Objective  Vector\\
			~ &  Dimension ($Dim$) & Dimension ($M$)\\
			\midrule
			ZDT1-3 & 30 & 2 \\
			\midrule
			ZDT4 & 10 & 2 \\
			\midrule
			ZDT5 & 11 & 2 \\
			\midrule
			ZDT6 & 10 & 2 \\
			\midrule
			DTLZ1-7 & 11 & 2 \\
			\midrule
			WFG1-9 & 12 & 3 \\
			\midrule
			UF1-7 & 30 & 2 \\
			\midrule
			UF8-10 & 30 & 3 \\
			\midrule
			MaOP1-10 & 10 & 3\\ 
			\bottomrule
	\end{tabular}}
\end{table}

\begin{table}[tbp]
	\centering
	\caption{The three foundation algorithms and their evolutionary operators and parameters that define the algorithm configuration space. $Dim$ represents the decision vector dimension of the problem.}
		\renewcommand{\arraystretch}{0.68}
	\scalebox{0.8}{
		\begin{tabular}{@{}llll@{}}
			\toprule
			\multicolumn{2}{c}{Foundation Algorithm}& Parameter & Value Range \\
			\midrule
			\multirow{27}{*}{NSGA-II} &\multirow{5}{*}{\textit{\textbf{SBX+PM}}} & $\eta(\textit{\textbf{SBX}})$ & $\{1, 2, \ldots, 100\}$ \\
			\cmidrule{3-4}
			~ &~ & $p_{c}$ & $\{1\}$\\
			\cmidrule{3-4}
			~ &~ & $\eta(\textit{\textbf{PM}}) $& $ \{1, 2, \ldots,100$\}\\
			\cmidrule{3-4}
			~ &~ & $p_{m}$ & $\{1/Dim\}$\\
			\cmidrule{2-4}
			~ &\multirow{4}{*}{\textit{\textbf{rand/p}}} & $F $&$ \left(0,2\right]$\\
			\cmidrule{3-4}
			~ &~& $p $&$ \{1, 2\}$\\
			\cmidrule{3-4}
			~ &~& $CR $&$ \left(0,1\right]$\\
			\cmidrule{2-4}
			~ &\multirow{4}{*}{\textit{\textbf{best/p}}} & $F $&$ \left(0,2\right]$\\
			\cmidrule{3-4}
			~ &~& $p $&$ \{1, 2\}$\\
			\cmidrule{3-4}
			~ &~& $CR $&$ \left(0,1\right]$\\
			\cmidrule{2-4}
			~ &\multirow{3}{*}{\textit{\textbf{current-to-}}} & $F $&$ \left(0,2\right]$\\
			\cmidrule{3-4}
			~ &~& $K $&$ \left(0,1\right]$\\
			\cmidrule{3-4}
			~ &\multirow{3}{*}{\textit{\textbf{rand/p}}}& $p $&$ \{1\}$\\
			\cmidrule{3-4}
			~ &~& $CR $&$ \left(0,1\right]$\\
			\cmidrule{2-4}
			~ &\multirow{3}{*}{\textit{\textbf{current-to-}}} & $F $&$ \left(0,2\right]$\\
			\cmidrule{3-4}
			~ &~& $K $&$ \left(0,1\right]$\\
			\cmidrule{3-4}
			~ &\multirow{3}{*}{\textit{\textbf{best/p}}}& $p $&$ \{1\}$\\
			\cmidrule{3-4}
			~ &~& $CR $&$ \left(0,1\right]$\\
			\midrule
			\multirow{16}{*}{MOEA/D} &\multirow{6}{*}{\textit{\textbf{SBX+PM}}} & $\eta(\textit{\textbf{SBX}})$ & $\{1, 2, \ldots, 100\}$ \\
			\cmidrule{3-4}
			~ &~ & $p_{c}$ & $\{1\}$\\
			\cmidrule{3-4}
			~ &~ & $\eta(\textit{\textbf{PM}}) $& $ \{1, 2, \ldots,100$\}\\
			\cmidrule{3-4}
			~ &~ & $p_{m}$ & $\{1/Dim\}$\\
			\cmidrule{2-4}
			~ &\multirow{4}{*}{\textit{\textbf{rand/p}}} & $F $&$ \left(0,2\right]$\\
			\cmidrule{3-4}
			~ &~& $p $&$ \{1, 2\}$\\
			\cmidrule{3-4}
			~ &~& $CR $&$ \left(0,1\right]$\\
			\cmidrule{2-4}
			~ &\multirow{3}{*}{\textit{\textbf{current-to-}}} & $F $&$ \left(0,2\right]$\\
			\cmidrule{3-4}
			~ &~& $K $&$ \left(0,1\right]$\\
			\cmidrule{3-4}
			~ &\multirow{3}{*}{\textit{\textbf{rand/p}}}& $p $&$ \{1\}$\\
			\cmidrule{3-4}
			~ &~& $CR $&$ \left(0,1\right]$\\
			\cmidrule{2-4}
			~ & ~& $ P_s $ & $\left[0,1\right]$\\
			\cmidrule{3-4}
			~ & ~& $ n_r $ & $\{ 2, 3,\ldots,10 \}$\\
			\cmidrule{3-4}
			~ & ~& $ neighborSize $ & $\{ 10, 11,\ldots ,50 \}$\\			
			\midrule
			\multirow{15}{*}{MOPSO} & \multirow{2}{*}{SMPSO} & \textit{\textbf{PMn}} &  $ \{1, 2, \ldots,100$\} \\
			\cmidrule{3-4}
			~ & ~ & $Constrition$ &  $ \{ True, False \}$ \\
			\cmidrule{2-4}
			~& OMOPSO & \textit{\textbf{b}} &  $ \{1, 2, \ldots,20$ \} \\
			\cmidrule{2-4}
			~ &  \multicolumn{3}{c}{NO Mutate} \\
			\cmidrule{2-4}
			~ & ~ & \textit{\textbf{w}} & $ \left[0,1\right]$ \\
			\cmidrule{3-4}
			~ & ~ & \textit{\textbf{C1}} & $ \left[0.5,2.5\right]$ \\
			\cmidrule{3-4}
			~ & ~ & \textit{\textbf{C2}} & $ \left[0.5,2.5\right]$ \\
			\cmidrule{3-4}
			~ & ~ &  $ V_{max} $ & $ \left[0.5,10\right]$ \\
			\cmidrule{3-4}
			~ & ~ & \textit{\textbf{M}} & $ \left[5,20\right]$ \\
			\cmidrule{3-4}
			~ & ~ & $ V_{change}$ & $ \{ 1, 0.1, 0.01, 0.001, -1\} $ \\			
			\bottomrule
	\end{tabular}}
	\label{operator parameter}
\end{table}

\section{Experiments}
\label{experiments}
In the experiments, we first used the proposed automatic construction approach to build a MOEAs/PAP with a training set, and then compared it with other MOEAs on an independent testing set to verify its performance. 
Besides, we compared it with the three foundation algorithms on the testing set.
All compared algorithms were fine-tuned on the training set in the same way as Algorithm~\ref{automatic construction algorithm}.
Finally, we analyzed the performance of each member algorithm of the MOEAs/PAP, as well as the impact of the \textbf{Restructure} procedure.

\subsection{Benchmark Sets}
\label{exp_instance_sets}
We collected the commonly-used MOP benchmarks in the literature, including ZDT~\citep{num16}, DTLZ~\citep{num17}, WFG~\citep{num18}, UF~\citep{num19} and MaOP~\citep{MaOP}.
These benchmark sets contain 42 MOPs in total, as summarized in Table~\ref{problem parameter}.
From each of these benchmark set, several problems were chosen randomly as the training problems, and the remaining problems were used for testing.
	The number of training problems and testing problems are basically equal.
Specifically, the training set includes 21 MOPs in total, i.e., UF2-3, UF5, UF9--10, WFG2-4, WFG6, WFG9, DTLZ3-4, DTLZ6, ZDT3-5, MaOP5, MaOP7, MaOP7-8, and MaOP10; the testing set contains 21 MOPs in total, i.e., UF1, UF4, UF6-8, WFG1, WFG5, WFG7-8, DTLZ1-2, DTLZ5, DTLZ7, DTLZ1, DTLZ6-7, ZDT1-2, ZDT6, MaOP1, MaOP3-4, MaOP6 and MaOP9.

\begin{table}[tbp]
	\centering
	\caption{Details of the member algorithms of the constructed MOEAs/PAP.}
	\scalebox{0.9}{
		\begin{tabular}{@{}lll@{}}
			\toprule
			\multicolumn{2}{c}{Member Algorithm} & Parameter \\
			\midrule
			\multirow{5}{*}{\textit{\textbf{MOEA/D}}} & \multirow{2}{*}{\textit{\textbf{SBX+PM}}} & $\eta(\textit{\textbf{SBX}}) = 1, \eta(\textit{\textbf{PM}}) = 48$\\
			~ &~ & $p_c = 1, p_m = 1/Dim$\\
			\cmidrule{2-3}
			~ &~ & $P_s = 0.903, n_r = 9$\\
			\cmidrule{3-3}
			~ &~ & $neighborSize = 50$\\
			\midrule
			\multirow{2}{*}{\textit{\textbf{NSGA-II}}} & \multirow{2}{*}{\textit{\textbf{rand/p}}} & $p = 1, F = 1.072$\\
			~ &~ & $CR = 0.026$\\
			\midrule
			\multirow{5}{*}{\textit{\textbf{MOEA/D}}} & \multirow{2}{*}{\textit{\textbf{SBX+PM}}} & $\eta(\textit{\textbf{SBX}}) = 62, \eta(\textit{\textbf{PM}}) = 5$\\
			~ &~ & $p_c = 1, p_m = 1/Dim$\\
			\cmidrule{2-3}
			~ &~ & $P_s = 0.794, n_r = 9$\\
			\cmidrule{3-3}
			~ &~ & $neighborSize = 29$\\
			\midrule
			\multirow{2}{*}{\textit{\textbf{NSGA-II}}} & \multirow{2}{*}{\textit{\textbf{rand/p}}} & $p = 1, F = 0.136$\\
			~ &~ & $CR = 0.681$\\
			\midrule
			\multirow{5}{*}{\textit{\textbf{MOEA/D}}} & \multirow{2}{*}{\textit{\textbf{rand/p}}} & $p = 1, F = 0.753$\\
			~ &~ & $CR=0.963$\\
			\cmidrule{2-3}
			~ &~ & $P_s = 0.645, n_r = 3$\\
			\cmidrule{3-3}
			~ &~ & $neighborSize = 42$\\
			\midrule
			\multirow{5}{*}{\textit{\textbf{MOEA/D}}} & \multirow{2}{*}{\textit{\textbf{SBX+PM}}} & $\eta(\textit{\textbf{SBX}}) = 89, \eta(\textit{\textbf{PM}}) = 2$\\
			~ &~ & $p_c = 1, p_m = 1/Dim$\\
			\cmidrule{2-3}
			~ &~ & $P_s = 0.303, n_r = 2$\\
			\cmidrule{3-3}
			~ &~ & $neighborSize = 38$\\
			\bottomrule
	\end{tabular}}
	\label{parameter for pap}
\end{table}

\subsection{Construction of MOEAs/PAP}
\label{exp_automatic_construction}
To use the approach in Algorithm~\ref{automatic construction algorithm} to construct a MOEAs/PAP, one needs to provide a training set of MOPs (detailed in Section~\ref{exp_instance_sets}) and an algorithm configuration space.
Here, the algorithm configuration space is defined based on three foundation algorithms, NSGA-II, MOEA/D and MOPSO.
All the foundation algorithms, as well as their parameters, are summarized in Table~\ref{operator parameter}.
In the tablem, $V_{max} $ represents the ratio of the maximum value of the velocity to the range of values of the decision variable in the problem. And $V_{change}$ represents the parameter taken when the correction is made after the speed reaches the maximum.
Finally, the maximum number of member algorithms, i.e., $k$, was set to 10, considering that 10-core machines are widely available now.

Given the training set and the algorithm configuration space, we used the proposed approach to construct a MOEAs/PAP.
Specifically, MOEAs/PAP contains six member algorithms, which are detailed in Table~\ref{parameter for pap}.

\begin{table}[tbp]
	\centering
	\caption{Parameter values of the compared  ensemble MOEAs after fine-tuning.}
	\scalebox{0.8}{
		\begin{tabular}{@{}lll@{}}
				\toprule
				Algorithm & Operator & Parameter \\
				\midrule
				\multirow{5}{*}{NSGA-II/MOE} & \textit{\textbf{rand/1}} & $F = 1.651, CR = 0.030$ \\
				\cmidrule{2-3}
				~ & \textit{\textbf{rand/2}} & $F = 0.807, CR = 0.005$\\
				\cmidrule{2-3}
				~ & \textit{\textbf{current-to-}} &  $F = 1.718, CR = 0.416$\\
				~ & \textit{\textbf{rand/1}} &  $ K=0.898$\\
				\cmidrule{2-3}
				~&  \multirow{2}{*}{\textit{\textbf{SBX+PM}}} & $\eta(\textit{\textbf{SBX}}) = 13, \eta(\textit{\textbf{PM}}) = 87$\\
				~ & ~ & $p_c = 1, p_m = 1/Dim$\\
				\midrule
				\multirow{12}{*}{MOEA/D-TAOS} & \textit{\textbf{current-to-}} & $F = 1.640$ \\
				~ & \textit{\textbf{rand/1}} & $ CR = 0.696$ \\
				\cmidrule{2-3}
				~ & \textit{\textbf{rand/1}} & $F = 0.030, CR = 0.559$\\
				\cmidrule{2-3}
				~ & \textit{\textbf{current-to-}} &  $F = 0.504$\\
				~ & \textit{\textbf{rand/1}} &  $ CR = 0.259$\\
				\cmidrule{2-3}
				~ & \textit{\textbf{rand/1}} & $F = 0.111, CR = 0.200$\\
				\cmidrule{2-3}
				~ & ~ & $\eta(\textit{\textbf{PM}}) = 26$\\
				\cmidrule{3-3}
				~ & ~ & $ p_m = 1/Dim $\\
				\cmidrule{3-3}
				~ & ~ & $ P_s = 0.788 $\\
				\cmidrule{3-3}
				~ & ~ & $ k = 1 $\\
				\cmidrule{3-3}
				~ & ~ & $ neighborSize = 27 $\\
				\cmidrule{3-3}
				~ & ~ & $ n_r = 8 $\\
				\midrule
				\multirow{10}{*}{EF-PD} & \multirow{2}{*}{\textit{\textbf{rand/1 + PM}}} & $F = 0.436, CR = 0.917$ \\
				~ & ~ & $\eta(\textit{\textbf{PM}}) = 33, p_m = 1/n$\\
				\cmidrule{2-3}
				~&  \multirow{2}{*}{\textit{\textbf{SBX+PM}}} & $\eta(\textit{\textbf{SBX}}) = 90, \eta(\textit{\textbf{PM}}) = 36$\\
				~ & ~ & $p_c = 1, p_m = 1/Dim$\\
				\cmidrule{2-3}
				~ & ~ & $ P_s = 0.062 $\\
				\cmidrule{3-3}
				~ & ~ & $ neighborSize = 18 $\\
				\cmidrule{3-3}
				~ & ~ & $ n_r = 8 $\\
				\bottomrule
			\end{tabular}}
	\label{algorithm configuration}
\end{table}

\begin{table}[tbp]
	\centering
	\caption{Parameter values of the compared  foundation algorithms after fine-tuning.}
		\begin{tabular}{@{}lll@{}}
			\toprule
			\multicolumn{2}{c}{Foundation Algorithm} & Parameter \\
			\midrule
			\multirow{5}{*}{\textit{\textbf{MOEA/D}}} & \multirow{2}{*}{\textit{\textbf{SBX+PM}}} & $\eta(\textit{\textbf{SBX}}) = 26, \eta(\textit{\textbf{PM}}) = 75$\\
			~ &~ & $p_c = 1, p_m = 1/Dim$\\
			\cmidrule{2-3}
			~ &~ & $P_s = 0.879, n_r = 10$\\
			\cmidrule{3-3}
			~ &~ & $neighborSize = 50$\\
			\midrule
			\multirow{2}{*}{\textit{\textbf{NSGA-II}}} & \multirow{2}{*}{\textit{\textbf{rand/p}}} & $p = 2, F = 0.224$\\
			~ &~ & $CR = 0.372$\\
			\midrule
			\multirow{5}{*}{\textit{\textbf{MOPSO}}} & OMOPSO & $ b = 8 $\\
			\cmidrule{2-3}
			~ &~ & $w = 0.075, V_{max} = 3.794$\\
			\cmidrule{3-3}
			~ &~ & $C1 = 1.985, C2 = 1.560$\\
			\cmidrule{3-3}
			~ &~ & $V_{change} = 0.01$\\
			\bottomrule
	\end{tabular}
	\label{foundation algorithm configuration}
\end{table}

\subsection{Compared Algorithms and Experimental Protocol}
We compared MOEAs/PAP with the state-of-the-art ensemble MOEAs, NSGA-II/MOE \citep{num6}, EF-PD \citep{EFPD} and MOEA/D-TAOS \citep{TAOS}. Besides, the foundation algorithms, NSGA-II, MOEA/D and MOPSO, are also compared.
	For the compared algorithms, the recommended parameter settings are fine-tuned by Alogrithm~\ref{automatic construction algorithm} as well, which are listed in Table \ref{algorithm configuration} and Table \ref{foundation algorithm configuration} respectively.

\begin{sidewaystable*}
	\centering
	\caption{The mean $\pm$ variance of the HV values across the 30 runs over the testing set.
			For each testing problem, the best performance is indicated in bold (note for HV, the larger, the better).
			$\dagger$ indicates that the performance of the algorithm is not significantly different from the performance of MOEAs/PAP (according to a Wilcoxon?s bilateral rank sum test with $p=0.05$.
			The significance test of MOEAs/PAP against the ensemble MOEAs is summarized in the win-draw-loss (W-D-L) counts.}
	\scalebox{.75}{
		\begin{tabular}{@{}lccccccc@{}}
			\toprule
			Problem& MOEAs/PAP        & EF-PD(\textbf{\textit{Nsize}})        & EF-PD(\textbf{\textit{Ngen}})        & MOEA/D-TAOS(\textbf{\textit{Nsize}})        & MOEA/D-TAOS(\textbf{\textit{Ngen}})        & NSGA-II/MOE(\textbf{\textit{Nsize}})        & NSGA-II/MOE(\textbf{\textit{Ngen}})   \\
			\midrule
			UF1   & 0.6860  $\pm$ 1.46E-04 & \textbf{0.6939 } $\pm$ 5.50E-05 & 0.6513  $\pm$ 9.91E-03 & 0.6568 $\pm$ 2.98E-04 & 0.6744  $\pm$ 7.37E-05 & 0.6150  $\pm$ 4.06E-04 & 0.6324  $\pm$ 8.17E-04 \\
			\midrule
			UF4   & \textbf{0.4419 } $\pm$ 1.34E-07 & 0.4393  $\pm$ 1.45E-06 & 0.4397 $\pm$ 2.25E-06 & 0.4394  $\pm$2.05E-06 & 0.4404  $\pm$ 1.27E-06 & 0.4395  $\pm$ 3.12E-06 & 0.4407  $\pm$ 8.02E-08 \\
			\midrule
			UF6   & \textbf{0.6902 } $\pm$ 1.31E-03 & 0.5613  $\pm$ 5.87E-03 & 0.4185 $\pm$ 1.18E-02 & 0.6679  $\pm$ 1.43E-03 & 0.6070  $\pm$ 3.43E-03 & 0.6375  $\pm$ 5.30E-03 & 0.5962  $\pm$ 6.94E-03 \\
			\midrule
			UF7   & \textbf{0.5615 } $\pm$ 3.21E-05 & 0.5557  $\pm$ 4.79E-05 & 0.4781  $\pm$ 2.56E-02  & 0.5556  $\pm$ 1.38E-03$\dagger$ & 0.5460  $\pm$ 2.98E-03 & 0.5127  $\pm$2.06E-03 & 0.4889  $\pm$ 1.03E-02 \\
			\midrule
			UF8   & \textbf{0.4239 } $\pm$ 9.26E-04 & 0.3548  $\pm$ 1.30E-02 & 0.2805  $\pm$ 1.90E-02 & 0.3998  $\pm$ 2.15E-03 & 0.3478  $\pm$ 1.22E-03 & 0.2753  $\pm$ 3.07E-03 & 0.2849  $\pm$ 1.89E-03 \\
			\midrule
			WFG1  & 0.8183  $\pm$ 5.49E-04 & 0.8377  $\pm$ 1.18E-03 & 0.9223  $\pm$ 8.99E-05 & 0.8697  $\pm$ 1.49E-04 & 0.9103  $\pm$ 1.27E-04 & 0.9270  $\pm$ 5.17E-05 & \textbf{0.9416 } $\pm$ 4.49E-06 \\
			\midrule
			WFG5  & 0.5055  $\pm$ 8.89E-06 & 0.4950 $\pm$ 2.53E-05 & 0.5022 $\pm$ 2.42E-05 & 0.4959  $\pm$ 1.72E-05 & 0.5028 $\pm$ 2.90E-05 & 0.4983  $\pm$ 2.77E-05 & \textbf{0.5073 }$\pm$ 1.45E-05 \\
			\midrule
			WFG7  & 0.0911  $\pm$ 2.98E-06 & 0.0758  $\pm$ 1.96E-05 & 0.0744 $\pm$ 2.02E-05 & 0.0832  $\pm$ 8.02E-06 & 0.0827 $\pm$ 4.22E-06 & 0.0874  $\pm$ 1.14E-05 & \textbf{0.0937 } $\pm$ 8.97E-06 \\
			\midrule
			WFG8  & 0.2291  $\pm$ 7.78E-07 & 0.2306 $\pm$ 5.41E-06 & 0.2335 $\pm$ 2.05E-06 & 0.2296  $\pm$5.20E-06 & 0.2308 $\pm$ 3.90E-06 & 0.2103 $\pm$ 1.04E-05 & \textbf{0.2421 } $\pm$ 8.21E-08 \\
			\midrule
			DTLZ1 & 0.4970 $\pm$ 3.49E-03 & 0.5577 $\pm$ 1.07E-02 & 0.5784 $\pm$1.49E-06 & 0.5772  $\pm$ 1.80E-06 & 0.5783 $\pm$ 8.76E-06 & 0.5781  $\pm$1.28E-06 & \textbf{0.5789 } $\pm$ 4.30E-06 \\
			\midrule
			DTLZ2 & \textbf{0.3468} $\pm$ 6.37E-08 & 0.3423 $\pm$2.36E-06 & 0.3441 $\pm$ 5.44E-07 & 0.3425  $\pm$ 1.83E-06 & 0.3438  $\pm$ 7.24E-07 & 0.3427  $\pm$ 2.51E-06 & 0.3466 $\pm$ 6.02E-08 \\
			\midrule
			DTLZ5 & \textbf{0.3467 }$\pm$ 6.21E-08 & 0.3419  $\pm$ 4.35E-06 & 0.3440  $\pm$ 1.11E-06 & 0.3427  $\pm$ 2.09E-06 & 0.3436 $\pm$ 8.13E-07 & 0.3430  $\pm$ 3.32E-06 & 0.3465 $\pm$ 9.74E-08 \\
			\midrule
			DTLZ7 & 0.2425 $\pm$ 9.51E-09 & 0.2413  $\pm$ 4.23E-07 & 0.2415 $\pm$ 2.74E-07 & 0.2406 $\pm$ 6.82E-07 & 0.2410 $\pm$ 1.04E-06 & 0.2412 $\pm$ 1.34E-06 & \textbf{0.2425 }$\pm$ 1.20E-08 \\
			\midrule
			ZDT1  & 0.7181 $\pm$ 3.80E-07 & 0.7154 $\pm$ 1.54E-06 & 0.7159 $\pm$ 1.78E-06 & 0.7147 $\pm$ 5.95E-06 & 0.7163 $\pm$ 1.43E-06 & 0.7173 $\pm$ 3.05E-07 & \textbf{0.7198 }$\pm$ 3.58E-08 \\
			\midrule
			ZDT2  & 0.4426 $\pm$ 2.08E-07 & 0.4394 $\pm$ 2.67E-06 & 0.4406 $\pm$ 5.07E-06 & 0.4400 $\pm$ 1.52E-06 & 0.4413 $\pm$ 1.66E-06 & 0.4394 $\pm$ 2.97E-06 & \textbf{0.4443 }$\pm$ 3.88E-08 \\
			\midrule
			ZDT6  & 0.3870 $\pm$ 1.09E-07 & 0.3849 $\pm$ 7.81E-07 & 0.3845 $\pm$ 1.40E-06 & 0.3853 $\pm$ 6.47E-07 & 0.3861 $\pm$ 6.00E-07 & 0.3851 $\pm$ 2.18E-06 & \textbf{0.3875 }$\pm$ 4.05E-07 \\
			\midrule
			MaOP1 & \textbf{0.2331 } $\pm$ 1.29E-06 & 0.0632 $\pm$1.75E-05 & 0.0625 $\pm$ 2.49E-05 & 0.0522 $\pm$ 1.59E-07 & 0.0524 $\pm$5.64E-08 & 0.2314 $\pm$ 5.35E-07 & 0.2315 $\pm$5.52E-07 \\
			\midrule
			MaOP3 & \textbf{0.4953 }$\pm$ 5.28E-08 & 0.0136 $\pm$ 6.66E-04 & 0.4928 $\pm$ 2.78E-07 & 0.4932$\pm$ 3.40E-07 & 0.4823  $\pm$ 3.13E-03 & 0.4921 $\pm$ 2.18E-06 & 0.4943$\pm$ 1.92E-06 \\
			\midrule
			MaOP4 & 0.4276 $\pm$ 9.77E-04 & 0.4281 $\pm$ 1.30E-03 & 0.4335 $\pm$ 1.43E-03 $\dagger$& 0.4312 $\pm$ 1.19E-03 $\dagger$ & \textbf{0.4548 } $\pm$ 2.30E-03 & 0.3598 $\pm$ 5.52E-04 & 0.4084 $\pm$ 5.96E-09 \\
			\midrule
			MaOP6 & \textbf{0.8221 }$\pm$ 2.28E-06 & 0.7522 $\pm$ 1.91E-03 & 0.7501 $\pm$ 1.87E-03 & 0.7491 $\pm$ 1.21E-03 & 0.7584 $\pm$ 8.78E-04 & 0.8044 $\pm$ 1.38E-04 & 0.8137 $\pm$ 9.88E-06 \\
			\midrule
			MaOP9 & \textbf{0.4307 }$\pm$ 9.76E-07 & 0.4291  $\pm$ 8.97E-07 & 0.4294 $\pm$ 1.81E-07 & 0.4269 $\pm$ 2.22E-06 & 0.4262 $\pm$ 3.61E-07 & 0.4197 $\pm$ 1.53E-04 & 0.4155  $\pm$5.35E-04 \\
			\midrule
			\midrule
			W-D-L  & - & 16-0-5 & 17-1-3 & 16-2-3 & 17-0-4 & 19-0-2 & 12-0-9 \\
			\bottomrule
		\end{tabular}%
	}
	\label{HV comparison}%
\end{sidewaystable*}%

% Table generated by Excel2LaTeX from sheet '2'
\begin{sidewaystable*}
	\centering
	\caption{The mean $\pm$ variance of the HV values across the 30 runs over the testing set.
			For each testing problem, the best performance is indicated in bold (note for HV, the larger, the better).
			$\dagger$ indicates that the performance of the algorithm is not significantly different from the performance of MOEAs/PAP (according to a Wilcoxon?s bilateral rank sum test with $p=0.05$.
			The significance test of MOEAs/PAP against the foundation algorithms is summarized in the win-draw-loss (W-D-L) counts.}
	\scalebox{.83}{
		\begin{tabular}{@{}lccccccc@{}}
			\toprule
			Problem& MOEAs/PAP        & MOEA/D(\textbf{\textit{Ngen}})       & MOEA/D(\textbf{\textit{Nsize}})        & NSGA-II(\textbf{\textit{Ngen}})        & NSGA-II(\textbf{\textit{Nsize}})        & MOPSO(\textbf{\textit{Ngen}})        & MOPSO(\textbf{\textit{Nsize}})   \\
			\midrule
			UF1   & \textbf{0.6860 } $\pm$ 1.46E-04 & 0.5741  $\pm$ 1.98E-02 & 0.5807  $\pm$ 1.40E-02 & 0.6147  $\pm$ 9.54E-04 & 0.6119  $\pm$ 1.04E-04 & 0.6210  $\pm$ 7.89E-05 & 0.6130  $\pm$ 1.02E-04 \\
			\midrule
			UF4   & \textbf{0.4419 } $\pm$ 1.34E-07 & 0.4396  $\pm$ 1.86E-06 & 0.4390  $\pm$ 1.65E-06 & 0.4401  $\pm$9.10E-08 & 0.4393 $\pm$ 4.55E-06 & 0.4374  $\pm$ 7.84E-07 & 0.4380 $\pm$ 2.64E-06 \\
			\midrule
			UF6   & \textbf{0.6902 } $\pm$ 1.31E-03 & 0.4108 $\pm$ 1.83E-02 & 0.4182 $\pm$  2.21E-02 & 0.5769 $\pm$  3.06E-03 & 0.5985 $\pm$  1.19E-02 & 0.6000 $\pm$  4.59E-03 & 0.5718 $\pm$  2.41E-03 \\
			\midrule
			UF7   & \textbf{0.5615 }$\pm$  3.21E-05 & 0.3177 $\pm$  3.28E-02 & 0.2983 $\pm$  3.66E-02 & 0.4897 $\pm$  1.15E-02 & 0.5192 $\pm$  2.64E-04 & 0.5453  $\pm$ 1.14E-05 & 0.5378 $\pm$  1.38E-05 \\
			\midrule
			UF8   & \textbf{0.4239 }$\pm$  9.26E-04 & 0.3411 $\pm$  3.75E-03 & 0.3527 $\pm$  5.25E-03 & 0.2689 $\pm$  3.56E-03 & 0.2260 $\pm$  8.51E-04 & 0.2913 $\pm$  3.08E-03 & 0.3496 $\pm$  1.59E-03 \\
			\midrule
			WFG1  & 0.8183  $\pm$  5.49E-04 & 0.8980 $\pm$  2.23E-04 & 0.8792 $\pm$  2.92E-04 & \textbf{0.9261 }$\pm$  3.97E-05 & 0.8921 $\pm$  2.42E-04 & 0.2224 $\pm$  9.21E-05 & 0.2304  $\pm$  2.29E-04 \\
			\midrule
			WFG5  & 0.5055 $\pm$  8.89E-06 & 0.5008 $\pm$  2.47E-05 & 0.4958 $\pm$  4.26E-05 & \textbf{0.5087 } $\pm$  1.11E-05 & 0.4985 $\pm$  1.48E-05 & 0.4543 $\pm$  1.11E-04 & 0.4883 $\pm$  2.36E-05 \\
			\midrule
			WFG7  & \textbf{0.0911 } $\pm$  2.98E-06 & 0.0811  $\pm$  4.47E-06 & 0.0828 $\pm$  5.66E-06 & 0.0854 $\pm$  3.58E-05 & 0.0423  $\pm$  1.62E-05 & 0.0110  $\pm$  1.59E-05 & 0.0107 $\pm$  3.01E-05 \\
			\midrule
			WFG8  & 0.2291 $\pm$  7.78E-07 & 0.2299 $\pm$  1.57E-06 & 0.2277 $\pm$  1.53E-06 & \textbf{0.2362 } $\pm$  5.15E-05 & 0.1157 $\pm$  8.85E-06 & 0.0440 $\pm$  3.31E-07 & 0.0504 $\pm$  8.94E-07 \\
			\midrule
			DTLZ1 & \textbf{0.4970 }$\pm$  3.49E-03 & 0.3496  $\pm$  7.63E-02 $\dagger$& 0.2090 $\pm$  7.12E-02 & 0.0000  $\pm$  0.00E+00 & 0.0383 $\pm$  2.05E-02 & 0.0000 $\pm$  0.00E+00 & 0.0000 $\pm$  0.00E+00 \\
			\midrule
			DTLZ2 & \textbf{0.3468 }$\pm$  6.37E-08 & 0.3439 $\pm$  1.01E-06 & 0.3425 $\pm$  2.02E-06 & 0.3459 $\pm$  6.40E-08 & 0.3421 $\pm$  6.62E-06 & 0.3445 $\pm$  1.05E-06 & 0.3398 $\pm$  7.10E-06 \\
			\midrule
			DTLZ5 & \textbf{0.3467 } $\pm$  6.21E-08 & 0.3439  $\pm$  1.01E-06 & 0.3424 $\pm$  2.00E-06 & 0.3459  $\pm$  6.64E-08 & 0.3423 $\pm$  7.06E-06 & 0.3445 $\pm$  1.05E-06 & 0.3397 $\pm$  7.03E-06 \\
			\midrule
			DTLZ7 & \textbf{0.2425 }$\pm$  9.51E-09 & 0.2412  $\pm$  1.46E-06 & 0.2402  $\pm$  2.90E-06 & 0.2398  $\pm$  1.96E-04 $\dagger$& 0.2410 $\pm$  1.42E-06 & 0.2415 $\pm$  6.23E-07 & 0.2410 $\pm$  6.39E-07 \\
			\midrule
			ZDT1  & \textbf{0.7181 }$\pm$  3.80E-07 & 0.7169  $\pm$  1.47E-06 & 0.7158  $\pm$  1.08E-06 & 0.7089 $\pm$  2.41E-04 & 0.7170 $\pm$  5.15E-07 & 0.7161  $\pm$  5.72E-07 & 0.7151 $\pm$  1.06E-06 \\
			\midrule
			ZDT2  & \textbf{0.4426 }$\pm$  2.08E-07 & 0.4409 $\pm$  2.75E-06 & 0.4399 $\pm$  1.64E-06 & 0.0697 $\pm$  1.92E-02 & 0.4402  $\pm$  1.75E-06 & 0.4408 $\pm$  1.59E-06 & 0.4388 $\pm$  2.94E-06 \\
			\midrule
			ZDT6  & \textbf{0.3870 } $\pm$  1.09E-07 & 0.3858 $\pm$  6.25E-07 & 0.3850  $\pm$  8.61E-07 & 0.3353  $\pm$  1.73E-02 $\dagger$& 0.3846  $\pm$  2.66E-06 & 0.3853  $\pm$  7.20E-07 & 0.3849  $\pm$  9.28E-07 \\
			\midrule
			MaOP1 & \textbf{0.2331 } $\pm$  1.29E-06 & 0.0525 $\pm$  4.37E-08 & 0.0520 $\pm$  2.52E-07 & 0.2319 $\pm$  1.95E-06 & 0.2308 $\pm$  4.21E-06 & 0.1301 $\pm$  1.38E-03 & 0.0926 $\pm$  7.11E-05 \\
			\midrule
			MaOP3 & \textbf{0.4953 }$\pm$  5.28E-08 & 0.4926 $\pm$  2.84E-07 & 0.4934 $\pm$  5.48E-07 & 0.4841 $\pm$  3.09E-03 & 0.4914 $\pm$  5.03E-06 & 0.0003 $\pm$  1.11E-06 & 0.0000 $\pm$  0.00E+00 \\
			\midrule
			MaOP4 & 0.4276 $\pm$  9.77E-04 & \textbf{0.4299 } $\pm$  1.15E-03$\dagger$ & 0.4296  $\pm$  1.13E-03 $\dagger$& 0.4083 $\pm$  3.05E-09 & 0.3621 $\pm$  6.30E-04 & 0.4078 $\pm$  1.16E-08 & 0.4077 $\pm$  3.32E-08 \\
			\midrule
			MaOP6 & 0.8221 $\pm$  2.28E-06 & 0.7115 $\pm$  1.01E-04 & 0.7168 $\pm$  1.33E-04 & 0.8197  $\pm$  2.00E-06 & 0.8074  $\pm$  3.82E-05 & \textbf{0.8284 }$\pm$  1.38E-06 & 0.8276 $\pm$  5.54E-06 \\
			\midrule
			MaOP9 & 0.4307 $\pm$  9.76E-07 & 0.3825 $\pm$  7.98E-03 & 0.3921 $\pm$  5.50E-03 & \textbf{0.4317 } $\pm$  1.68E-07 & 0.4258 $\pm$  3.45E-05 & 0.4242 $\pm$  1.25E-06 & 0.4198 $\pm$  3.86E-06 \\
			\midrule
			\midrule
			W-D-L  & - & 17-2-2 & 19-1-1 & 15-2-4 & 20-0-1 & 20-0-1 & 20-0-1 \\
			\bottomrule
		\end{tabular}%
	}
	\label{HV comparison_basic}%
\end{sidewaystable*}%

% Table generated by Excel2LaTeX from sheet 'sheet 1'
\begin{table*}[t]
	\centering
	\caption{The mean of the IHVR values achieved by the member algorithms of MOEAs/PAP, the variant of MOEAs/PAP without the  \textit{\textbf{Restructure}} procedure, and  MOEAs/PAP, across 30 independent runs over the training set.
			For each problem, the best performance is indicated  in bold (note for IHVR, the larger, the better), and the best performance achieved among the member algorithms is indicated by underline.}
	\scalebox{.75}{
		\begin{tabular}{@{}lcccccccc@{}}
			\toprule
			\multirow{2}{*}{Problem}&	Member &	Member&	Member & Member &	Member & Member &	\multirow{2}{*}{No \textit{\textbf{Restructure}}}&	\multirow{2}{*}{MOEAs/PAP}\\
			~ & Algorithm 1 & Algorithm 2 & Algorithm 3 & Algorithm 4 & Algorithm 5 & Algorithm 6 &~ &~ \\
			\midrule
			UF2   & 0.8736  & \underline{0.9036}  & 0.8600  & 0.8511  & 0.8449  & 0.8383  & 0.9053  & \textbf{0.9279 } \\
			\midrule
			UF3   & 0.4705  & 0.5285  & 0.4913  & 0.5381  & \underline{0.5694}  & 0.4975  & 0.5859  & \textbf{0.6849 } \\
			\midrule
			UF5   & 0.5008  & 0.5028  & 0.5574  & 0.5027  & 0.5025  & \underline{0.5575}  & 0.5726  & \textbf{0.5879 } \\
			\midrule
			UF9   & 0.2789  & \underline{0.3845}  & 0.2740  & 0.2575  & 0.2841  & 0.2519  & 0.4482  & \textbf{0.5116 } \\
			\midrule
			UF10  & 0.4290  & 0.3658  & \underline{0.4695}  & 0.2903  & 0.4052  & 0.4157  & 0.5196  & \textbf{0.5260 } \\
			\midrule
			WFG2  & 0.5388  & 0.6927  & 0.5177  & \underline{0.7568}  & 0.3273  & 0.4687  & \textbf{0.7573 } & \textbf{0.7573 } \\
			\midrule
			WFG3  & 0.6878  & 0.7077  & 0.6827  & \underline{\textbf{0.7694 }} & 0.4200  & 0.6189  & \textbf{0.7694 } & \textbf{0.7694 } \\
			\midrule
			WFG4  & 0.8150  & \underline{\textbf{0.8854 }} & 0.8516  & 0.8191  & 0.6801  & 0.8244  & \textbf{0.8854 } & \textbf{0.8854 } \\
			\midrule
			WFG6  & 0.7524  & \underline{0.8272}  & 0.7683  & 0.7968  & 0.6924  & 0.7346  & 0.8300  & \textbf{0.8302 } \\
			\midrule
			WFG9  & 0.6506  & 0.5675  & 0.6508  & \underline{0.6792}  & 0.6481  & 0.6472  & 0.6792  & \textbf{0.6910 } \\
			\midrule
			DTLZ3 & 0.7854  & 0.7857  & \underline{0.8886}  & 0.7853  & 0.7851  & 0.8156  & 0.8964  & \textbf{0.8967 } \\
			\midrule
			DTLZ4 & 0.9764  & 0.9290  & 0.9696  & 0.9651  & 0.9884  & \underline{0.9896}  & 0.9928  & \textbf{0.9931 } \\
			\midrule
			DTLZ6 & 0.9899  & \underline{0.9922}  & 0.9886  & 0.9716  & 0.9766  & 0.9894  & 0.9925  & \textbf{0.9934 } \\
			\midrule
			ZDT3  & 0.9880  & \underline{0.9944}  & 0.9886  & 0.9153  & 0.9534  & 0.9875  & 0.9945  & \textbf{0.9946 } \\
			\midrule
			ZDT4  & \underline{0.9390}  & 0.3334  & 0.9030  & 0.3572  & 0.3334  & 0.7819  & 0.9637  & \textbf{0.9651 } \\
			\midrule
			ZDT5  & 0.9450  & \underline{\textbf{1.0000 }} & 0.9643  & \underline{\textbf{1.0000 }} & 0.9526  & 0.9635  & \textbf{1.0000 } & \textbf{1.0000 } \\
			\midrule
			MaOP2 & 0.3249  & 0.2628  & 0.2380  & 0.3266  & \underline{0.4565}  & 0.1900  & 0.4846  & \textbf{0.5790 } \\
			\midrule
			MaOP5 & 0.7855  & 0.8961  & 0.6814  & \underline{0.9160}  & 0.9157  & 0.6564  & 0.9443  & \textbf{0.9490 } \\
			\midrule
			MaOP7 & 0.7794  & 0.6342  & 0.7807  & \underline{0.7838}  & 0.6199  & 0.7543  & \textbf{0.8420 } & \textbf{0.8420 } \\
			\midrule
			MaOP8 & 0.7368  & 0.6008  & 0.7761  & 0.6653  & 0.5102  & \underline{0.7941}  & \textbf{0.8297 } & \textbf{0.8297 } \\
			\midrule
			MaOP10 & 0.7856  & 0.7939  & 0.7582  & \underline{0.9872}  & 0.9349  & 0.7983  & 0.9918  & \textbf{0.9958 } \\
			\bottomrule
		\end{tabular}%
	}
	\label{IHVR_all}%
\end{table*}%

For all the tested algorithms, the population sizes and the number of fitness evaluation (FEs) were kept the same.
Specifically, for UF1-7, population size was set to 100 and max generation number was set to 500;
	for UF8-10, population size was set to 150 and max generation number was set to 600;
	for WFG1-9, population size was set to 150 and max generation number was set to 250;
	for all DTLZ and ZDT, population size was set to 100 and max generation number was set to 250;
	for all MaOP, population size was set to 300 and max generation number was set to 500.

Since MOEAs/PAP runs its member algorithms in parallel, it would perform more FEs than the compared algorithms.
Hence, in the experiments, for each compared algorithm, we also considered two variants of it that performed the same number of FEs as MOEAs/PAP.
	Let $N$ be the number of member algorithms of MOEAs/PAP (here $N$=6).
	Specifically, the first variant, indicated by \textit{\textbf{Ngen}}, increased the maximum generation number of the algorithm to $N$ times the original number.
	The second variant, indicated by \textit{\textbf{Nsize}}, increased the population size to $N$ times the original population size, and used the \textit{\textbf{Restructure}} procedure (see Section~\ref{MOEAs/PAP}) to choose the final Pareto-optimal solution set of the original population size from the final population.
We implemented all the tested algorithms with Python (version 3.8) library Geatpy (version 2.6.0 \footnote{\url{https://github.com/geatpy-dev/geatpy}}).
All the experiments were conducted on an Intel Xeon machine with 96 GB RAM and 16 cores (2.10 GHz, 20 MB Cache), running Ubuntu 16.04.

For each testing problem, each tested algorithm would be independently applied for 30 times; the average and variance of the achieved  HV and IGD would be reported. 

\subsection{Testing Results and Analysis}
The testing results for ensemble MOEAs in terms of HV and IGD are reported in Table~\ref{HV comparison} and Table~\ref{IGD comparison}, respectively.
For each problem, the best performance is indicated in bold, and a Wilcoxon?s bilateral rank-sum test was conducted to judge whether the difference between the performance of the compared algorithm and the performance of MOEAs/PAP was significant (with $p$=0.05).

Overall, MOEAs/PAP achieves the best performance in both Table~\ref{HV comparison} and Table~\ref{IGD comparison}.
Based on the win-draw-loss (W-D-L) counts, MOEAs/PAP shows significant advantages over all the other algorithms. 

In terms of HV (Table~\ref{HV comparison}), MOEAs/PAP achieves the best performance on 10 out of 21 testing problems, which is performing best among all ensemble MOEAs. Specially, compared with each algorithm individually, it is obvious that MOEAs/PAP is the superior one.
A similar observation can also be made regarding IGD from Table~\ref{IGD comparison}. Although MOEAs/PAP and NSGA-II/MOE($Ngen$) both achieved the best performance on 8 out of 21 among all algorithms, in the comparison of them, MOEAs/PAP achieved a clear advantage, performing better in 10 problems and no significant difference from NSGA-II/MOE($Ngen$) in 5 problems.

Afterwards, compared with the foundation algorithms, the advantages of MOEAs/PAP are more obvious. In terms of HV (Table~\ref{HV comparison_basic}), MOEAs/PAP achieves the best performance on 15 out of 21 testing problems, that is far more than the foundation algorithms. Compared with the best foundation algorithm, MOEAs/PAP achieves a performance of 15 wins and 2 draws. We can see that the performance of the MOEAs/PAP framework is very obvious beyond the foundation algorithms.
	In terms of IGD (Table~\ref{IGD comparison basic}), We can draw the same conclusion. MOEAs/PAP achieves the best performance on 13 out of 21 testing problems, more than the best foundation algorithm NSGA-II($Ngen$) performing best on 4 problems.

It is worth mentioning that, although the performance metric used in the construction of MOEAs/PAP, i.e., IHVR (see Section~\ref{sec:metric}), is based on HV, MOEAs/PAP still performs strongly in terms of IGD on the testing problems, which indicates the effectiveness of IHVR.

In summary, MOEAs/PAP not only shows superior performance than conventional single-operator-based MOEAs such as NSGA-II, but also outperforms the human-designed ensemble MOEAs.

\subsection{Performances of Member Algorithms}
Table \ref{IHVR_all} reports the independent performance of each member algorithm of MOEAs/PAP, and  the performance of MOEAs/PAP, on the training problems in terms of the metric IHVR (see Section~\ref{sec:metric}).
The details of these six member algorithms can be found in Table~\ref{parameter for pap}.
As before, on each problem, the best performance is indicated in bold; moreover, the best performance achieved among the member algorithms is indicated by underline.
One can observe that the performance of different member algorithms vary a lot.
	For example, member algorithm 1 performed poorly on UF9 but achieved excellent performance on ZDT4, while member algorithm 2 performed exactly on the opposite.
	Actually, noticing that every member-algorithm column in Table~\ref{IHVR_all} has at least one cell underlined, this means each member algorithm achieved the best performance among all the member algorithms on at least one problem.
Leveraging such performance complementarity eventually leads to a powerful PAP, i.e., MOEAs/PAP, which performed better than any single member algorithm on all the problems.

\subsection{Effectiveness of the Restructure Procedure}
\label{Effect of Restructure}
To investigate on the effect of the \textbf{\textit{Restructure}} procedure, we removed it from MOEAs/PAP and tested the resulting PAP on the training set.
Note this PAP is a conventional PAP as defined in Eq.~\eqref{performance}, where the best solution set found by the member algorithms would be returned as its output.
The ``No \textbf{\textit{Restructure}}'' column in Table \ref{IHVR_all} presents the results.
It can be observed that due to the integration of  \textbf{\textit{Restructure}} procedure, MOEAs/PAP achieved performance improvement on 15 out of 21 problems, clearly verifying the effectiveness of the procedure.

\section{Conclusion}
\label{conclusion}
This work extended the realm of automatic construction of PAPs to multi-objective optimization.
Specifically, we proposed a variant of PAP, namely MOEAs/PAP, which involves a \textbf{\textit{Restructure}} procedure to better determine the output solution set of the PAP.
Then, we presented  an automatic construction approach for MOEAs/PAP which uses a novel metric to evaluate the performance of MOEAs/PAPs across multiple MOPs.
Based on a training set of MOPs and an algorithm configuration space defined by several foundation algorithms, our approach could build MOEAs/PAPs that outperformed existing single-operator based-MOEAs and the state-of-the-art ensemble MOEAs designed by human experts.

The promising results presented in this work have indicated the huge potential of the automatic construction of PAPs in the area of multi-objective optimization.
Further directions for investigations are listed below.
\begin{itemize}
	\item  The algorithm configuration space used in this work is still defined based on the general algorithm framework. In the literature, there have been some studies on developing highly-parameterized MOEA frameworks \citep{BezerraLS16,BezerraLS20}. It is valuable to apply our construction approach to these MOEA frameworks, hopefully leading to even better MOEAs/PAPs.
	\item	When constructing MOEAs/PAPs, it is important to maintain the diversity among the member algorithms. Hence, the population diversity preservation schemes  such as negatively correlated search \citep{TangYY16} can be introduced into the construction approach to promote cooperation between different member algorithms.
	\item In real-world applications, one may be unable to collect sufficient MOPs as training problems. How to automatically build powerful PAPs in these scenarios is also worth studying.
\end{itemize}

\bibliography{bibliography}% common bib file

\begin{thebibliography}{61}
\providecommand{\natexlab}[1]{#1}
\providecommand{\url}[1]{{#1}}
\providecommand{\urlprefix}{URL }
\providecommand{\doi}[1]{\url{https://doi.org/#1}}
\providecommand{\eprint}[2][]{\url{#2}}
 \bibcommenthead

\bibitem[{Asanovic et~al(2009)Asanovic, Bodik, Demmel, Keaveny, Keutzer,
  Kubiatowicz, Morgan, Patterson, Sen, Wawrzynek et~al}]{num15}
Asanovic K, Bodik R, Demmel J, et~al (2009) A view of the parallel computing
  landscape. Communications of the ACM 52(10):56--67

\bibitem[{Bezerra et~al(2016)Bezerra, L{\'{o}}pez{-}Ib{\'{a}}{\~{n}}ez, and
  St{\"{u}}tzle}]{BezerraLS16}
Bezerra LCT, L{\'{o}}pez{-}Ib{\'{a}}{\~{n}}ez M, St{\"{u}}tzle T (2016)
  Automatic component-wise design of multiobjective evolutionary algorithms.
  {IEEE} Transactions on Evolutionary Computation 20(3):403--417

\bibitem[{Bezerra et~al(2020)Bezerra, L{\'{o}}pez{-}Ib{\'{a}}{\~{n}}ez, and
  St{\"{u}}tzle}]{BezerraLS20}
Bezerra LCT, L{\'{o}}pez{-}Ib{\'{a}}{\~{n}}ez M, St{\"{u}}tzle T (2020)
  Automatically designing state-of-the-art multi- and many-objective
  evolutionary algorithms. Evolutionary Computation 28(2):195--226

\bibitem[{Bosman and Thierens(2003)}]{num21}
Bosman PA, Thierens D (2003) The balance between proximity and diversity in
  multiobjective evolutionary algorithms. IEEE transactions on evolutionary
  computation 7(2):174--188

\bibitem[{Coello et~al(2007)Coello, Lamont, Van~Veldhuizen
  et~al}]{coello2007evolutionary}
Coello CAC, Lamont GB, Van~Veldhuizen DA, et~al (2007) Evolutionary algorithms
  for solving multi-objective problems. Springer

\bibitem[{Coello and Lechuga(2002)}]{MOPSO}
Coello CC, Lechuga MS (2002) Mopso: A proposal for multiple objective particle
  swarm optimization. In: Proceedings of the 2002 Congress on Evolutionary
  Computation. CEC'02 (Cat. No. 02TH8600), IEEE, pp 1051--1056

\bibitem[{Das and Suganthan(2011)}]{DasS11}
Das S, Suganthan PN (2011) Differential evolution: {A} survey of the
  state-of-the-art. {IEEE} Transactions on Evolutionary Computation 15(1):4--31

\bibitem[{Deb(2000)}]{num24}
Deb K (2000) An efficient constraint handling method for genetic algorithms.
  Computer methods in applied mechanics and engineering 186(2-4):311--338

\bibitem[{Deb et~al(2002{\natexlab{a}})Deb, Pratap, Agarwal, and
  Meyarivan}]{num26}
Deb K, Pratap A, Agarwal S, et~al (2002{\natexlab{a}}) A fast and elitist
  multiobjective genetic algorithm: Nsga-ii. IEEE transactions on evolutionary
  computation 6(2):182--197

\bibitem[{Deb et~al(2002{\natexlab{b}})Deb, Thiele, Laumanns, and
  Zitzler}]{num17}
Deb K, Thiele L, Laumanns M, et~al (2002{\natexlab{b}}) Scalable
  multi-objective optimization test problems. In: Proceedings of the 2002
  Congress on Evolutionary Computation. CEC'02 (Cat. No. 02TH8600), IEEE, pp
  825--830

\bibitem[{Dong et~al(2022)Dong, Lin, Zhou, and Jiang}]{TAOS}
Dong L, Lin Q, Zhou Y, et~al (2022) Adaptive operator selection with
  test-and-apply structure for decomposition-based multi-objective
  optimization. Swarm and Evolutionary Computation 68:101013

\bibitem[{Elsayed et~al(2017)Elsayed, Sarker, and Coello}]{num8}
Elsayed S, Sarker R, Coello CAC (2017) Fuzzy rule-based design of evolutionary
  algorithm for optimization. IEEE Transactions on Cybernetics 49(1):301--314

\bibitem[{Emmerich et~al(2005)Emmerich, Beume, and Naujoks}]{num23}
Emmerich M, Beume N, Naujoks B (2005) An emo algorithm using the hypervolume
  measure as selection criterion. In: International Conference on Evolutionary
  Multi-Criterion Optimization, Springer, pp 62--76

\bibitem[{Freund and Schapire(1997)}]{FreundS97}
Freund Y, Schapire RE (1997) A decision-theoretic generalization of on-line
  learning and an application to boosting. Journal of Computer and System
  Sciences 55(1):119--139

\bibitem[{Gao et~al(2022)Gao, Liu, Tan, and Song}]{num7}
Gao X, Liu T, Tan L, et~al (2022) Multioperator search strategy for
  evolutionary multiobjective optimization. Swarm and Evolutionary Computation
  71:101073

\bibitem[{Gebser et~al(2007)Gebser, Kaufmann, Neumann, and
  Schaub}]{GebserKNS07a}
Gebser M, Kaufmann B, Neumann A, et~al (2007) \emph{clasp} : {A}
  conflict-driven answer set solver. In: Baral C, Brewka G, Schlipf JS (eds)
  Proceedings of the 9th International Conference on Logic Programming and
  Nonmonotonic Reasoning, {LPNMR}'2007, Tempe, AZ, pp 260--265

\bibitem[{Goh et~al(2010)Goh, Tan, Liu, and Chiam}]{GohTLC10}
Goh CK, Tan KC, Liu DS, et~al (2010) A competitive and cooperative
  co-evolutionary approach to multi-objective particle swarm optimization
  algorithm design. European Journal of Operational Research 202(1):42--54

\bibitem[{Hamadi and Wintersteiger(2013)}]{HamadiW13}
Hamadi Y, Wintersteiger CM (2013) Seven challenges in parallel {SAT} solving.
  {AI} Magazine 34(2):99--106

\bibitem[{Hong et~al(2018)Hong, Tang, Zhou, Ishibuchi, and Yao}]{num5}
Hong W, Tang K, Zhou A, et~al (2018) A scalable indicator-based evolutionary
  algorithm for large-scale multiobjective optimization. IEEE Transactions on
  Evolutionary Computation 23(3):525--537

\bibitem[{Hong et~al(2021)Hong, Yang, and Tang}]{num20}
Hong WJ, Yang P, Tang K (2021) Evolutionary computation for large-scale
  multi-objective optimization: A decade of progresses. International Journal
  of Automation and Computing 18(2):155--169

\bibitem[{Huband et~al(2006)Huband, Hingston, Barone, and While}]{num18}
Huband S, Hingston P, Barone L, et~al (2006) A review of multiobjective test
  problems and a scalable test problem toolkit. IEEE Transactions on
  Evolutionary Computation 10(5):477--506

\bibitem[{Khan and Zhang(2010)}]{DRA}
Khan W, Zhang Q (2010) Moea/d-dra with two crossover operators. In: 2010 UK
  workshop on computational intelligence (UKCI), IEEE, pp 1--6

\bibitem[{Li and Zhang(2008)}]{num27}
Li H, Zhang Q (2008) Multiobjective optimization problems with complicated
  pareto sets, moea/d and nsga-ii. IEEE transactions on evolutionary
  computation 13(2):284--302

\bibitem[{Li et~al(2019)Li, Deb, Zhang, Suganthan, and Chen}]{MaOP}
Li H, Deb K, Zhang Q, et~al (2019) Comparison between moea/d and nsga-iii on a
  set of novel many and multi-objective benchmark problems with challenging
  difficulties. Swarm and Evolutionary Computation 46:104--117.
  \doi{https://doi.org/10.1016/j.swevo.2019.02.003},
  \urlprefix\url{https://www.sciencedirect.com/science/article/pii/S2210650218307016}

\bibitem[{Lindauer et~al(2022)Lindauer, Eggensperger, Feurer, Biedenkapp, Deng,
  Benjamins, Ruhkopf, Sass, and Hutter}]{lindauer2022smac3}
Lindauer M, Eggensperger K, Feurer M, et~al (2022) Smac3: A versatile bayesian
  optimization package for hyperparameter optimization. Journal of Machine
  Learnng Research 23:54--1

\bibitem[{Liu et~al(2019)Liu, Tang, and Yao}]{num13}
Liu S, Tang K, Yao X (2019) Automatic construction of parallel portfolios via
  explicit instance grouping. In: Proceedings of the AAAI Conference on
  Artificial Intelligence, pp 1560--1567

\bibitem[{Liu et~al(2021)Liu, Tang, and Yao}]{LIU2021100927}
Liu S, Tang K, Yao X (2021) Memetic search for vehicle routing with
  simultaneous pickup-delivery and time windows. Swarm and Evolutionary
  Computation 66:100927

\bibitem[{Liu et~al(2022{\natexlab{a}})Liu, Peng, and Tang}]{abs-2211-12713}
Liu S, Peng F, Tang K (2022{\natexlab{a}}) Reliable robustness evaluation via
  automatically constructed attack ensembles. CoRR abs/2211.12713

\bibitem[{Liu et~al(2022{\natexlab{b}})Liu, Tang, and Yao}]{LiuTY22}
Liu S, Tang K, Yao X (2022{\natexlab{b}}) Generative adversarial construction
  of parallel portfolios. {IEEE} Transactions on Cybernetics 52(2):784--795

\bibitem[{Liu et~al(2022{\natexlab{c}})Liu, Yang, and Tang}]{LiuYT2022}
Liu S, Yang P, Tang K (2022{\natexlab{c}}) Approximately optimal construction
  of parallel algorithm portfolios by evolutionary intelligence. SCIENTIA
  SINICA Technologica 53(2):280--290

\bibitem[{Ma et~al(2015)Ma, Su, Simon, and Fei}]{EMBBO}
Ma H, Su S, Simon D, et~al (2015) Ensemble multi-objective biogeography-based
  optimization with application to automated warehouse scheduling. Engineering
  Applications of Artificial Intelligence 44:79--90

\bibitem[{Mezura-Montes et~al(2008)Mezura-Montes, Reyes-Sierra, and
  Coello}]{num25}
Mezura-Montes E, Reyes-Sierra M, Coello CAC (2008) Multi-objective optimization
  using differential evolution: a survey of the state-of-the-art. In: Advances
  in differential evolution. Springer, p 173--196

\bibitem[{Nebro et~al(2009)Nebro, Durillo, Garcia-Nieto, Coello, Luna, and
  Alba}]{SMPSO}
Nebro AJ, Durillo JJ, Garcia-Nieto J, et~al (2009) Smpso: A new pso-based
  metaheuristic for multi-objective optimization. In: 2009 IEEE Symposium on
  computational intelligence in multi-criteria decision-making (MCDM), IEEE, pp
  66--73

\bibitem[{Niu et~al(2021)Niu, Liu, Wang, Tan, and Li}]{WOS:000574099800001}
Niu B, Liu Q, Wang Z, et~al (2021) Multi-objective bacterial colony
  optimization algorithm for integrated container terminal scheduling problem.
  Natural Computing 20(1):89--104

\bibitem[{Parsopoulos et~al(2022)Parsopoulos, Tatsis, Kotsireas, and
  Pardalos}]{parsopoulos2022parallel}
Parsopoulos KE, Tatsis VA, Kotsireas IS, et~al (2022) Parallel algorithm
  portfolios with adaptive resource allocation strategy. Journal of Global
  Optimization pp 1--21

\bibitem[{Peng et~al(2010)Peng, Tang, Chen, and Yao}]{PengTCY10}
Peng F, Tang K, Chen G, et~al (2010) Population-based algorithm portfolios for
  numerical optimization. {IEEE} Transactions on Evolutionary Computation
  14(5):782--800

\bibitem[{Pereira et~al(2022)Pereira, Sousa, and Rocha}]{WOS:000582098300001}
Pereira V, Sousa P, Rocha M (2022) A comparison of multi-objective optimization
  algorithms for weight setting problems in traffic engineering. Natural
  Computing 21(3):507--522

\bibitem[{Qiu et~al(2020)Qiu, Zhu, Wu, Chen, Pedrycz, and Suganthan}]{VNEF}
Qiu W, Zhu J, Wu G, et~al (2020) Ensemble many-objective optimization algorithm
  based on voting mechanism. IEEE Transactions on Systems, Man, and
  Cybernetics: Systems 52(3):1716--1730

\bibitem[{Ralphs et~al(2018)Ralphs, Shinano, Berthold, and Koch}]{RalphsSBK18}
Ralphs TK, Shinano Y, Berthold T, et~al (2018) Parallel solvers for mixed
  integer linear optimization. In: Hamadi Y, Sais L (eds) Handbook of Parallel
  Constraint Reasoning. Springer, p 283--336

\bibitem[{de~Santiago~Junior et~al(2020)de~Santiago~Junior, {\"O}zcan, and
  de~Carvalho}]{HRISE}
de~Santiago~Junior VA, {\"O}zcan E, de~Carvalho VR (2020) Hyper-heuristics
  based on reinforcement learning, balanced heuristic selection and group
  decision acceptance. Applied Soft Computing 97:106760

\bibitem[{Sierra and Coello~Coello(2005)}]{OMOPSO}
Sierra MR, Coello~Coello CA (2005) Improving pso-based multi-objective
  optimization using crowding, mutation and $\epsilon$-dominance. In:
  Evolutionary Multi-Criterion Optimization: Third International Conference,
  EMO 2005, Guanajuato, Mexico, March 9-11, 2005. Proceedings 3, Springer, pp
  505--519

\bibitem[{Sun et~al(2021)Sun, Liu, B{\"a}ck, and Xu}]{num9}
Sun J, Liu X, B{\"a}ck T, et~al (2021) Learning adaptive differential evolution
  algorithm from optimization experiences by policy gradient. IEEE Transactions
  on Evolutionary Computation 25(4):666--680

\bibitem[{Tang et~al(2014)Tang, Peng, Chen, and Yao}]{TangPCY14}
Tang K, Peng F, Chen G, et~al (2014) Population-based algorithm portfolios with
  automated constituent algorithms selection. Information Science 279:94--104

\bibitem[{Tang et~al(2016)Tang, Yang, and Yao}]{TangYY16}
Tang K, Yang P, Yao X (2016) Negatively correlated search. {IEEE} Journal on
  Selected Areas in Communications 34(3):542--550

\bibitem[{Tang et~al(2021)Tang, Liu, Yang, and Yao}]{num14}
Tang K, Liu S, Yang P, et~al (2021) Few-shots parallel algorithm portfolio
  construction via co-evolution. IEEE Transactions on Evolutionary Computation
  25(3):595--607

\bibitem[{Tang et~al(2020)Tang, Li, Deng, Chen, Guo, and
  Huang}]{WOS:000585987200008}
Tang Q, Li Y, Deng Z, et~al (2020) Optimal shape design of an autonomous
  underwater vehicle based on multi-objective particle swarm optimization.
  Natural Computing 19(4):733--742

\bibitem[{Wang et~al(2020)Wang, Xu, Qiu, and Zhang}]{num6}
Wang C, Xu R, Qiu J, et~al (2020) Adaboost-inspired multi-operator ensemble
  strategy for multi-objective evolutionary algorithms. Neurocomputing
  384:243--255

\bibitem[{Wang et~al(2018)Wang, Yang, Lin, Zhang, Wong, Coello, and
  Chen}]{EFPD}
Wang W, Yang S, Lin Q, et~al (2018) An effective ensemble framework for
  multiobjective optimization. IEEE Transactions on Evolutionary Computation
  23(4):645--659

\bibitem[{Wang et~al(2019)Wang, Yang, Lin, Zhang, Wong, Coello, and
  Chen}]{WangYLZWCC19}
Wang W, Yang S, Lin Q, et~al (2019) An effective ensemble framework for
  multiobjective optimization. {IEEE} Transactions on Evolutionary Computation
  23(4):645--659

\bibitem[{Wu et~al(2019)Wu, Mallipeddi, and Suganthan}]{wu2019ensemble}
Wu G, Mallipeddi R, Suganthan PN (2019) Ensemble strategies for
  population-based optimization algorithms--a survey. Swarm and evolutionary
  computation 44:695--711

\bibitem[{Zhang and Li(2007{\natexlab{a}})}]{ZhangL07}
Zhang Q, Li H (2007{\natexlab{a}}) {MOEA/D:} {A} multiobjective evolutionary
  algorithm based on decomposition. {IEEE} Transactions on Evolutionary
  Computation 11(6):712--731

\bibitem[{Zhang and Li(2007{\natexlab{b}})}]{MOEAD}
Zhang Q, Li H (2007{\natexlab{b}}) Moea/d: A multiobjective evolutionary
  algorithm based on decomposition. IEEE Transactions on Evolutionary
  Computation 11(6):712--731. \doi{10.1109/TEVC.2007.892759}

\bibitem[{Zhang et~al(2008)Zhang, Zhou, Zhao, Suganthan, Liu, Tiwari
  et~al}]{num19}
Zhang Q, Zhou A, Zhao S, et~al (2008) Multiobjective optimization test
  instances for the cec 2009 special session and competition. University of
  Essex, Colchester, UK and Nanyang technological University, Singapore,
  special session on performance assessment of multi-objective optimization
  algorithms, technical report 264:1--30

\bibitem[{Zhang et~al(2020)Zhang, Ren, Li, and Xuan}]{PAPHH}
Zhang S, Ren Z, Li C, et~al (2020) A perturbation adaptive pursuit strategy
  based hyper-heuristic for multi-objective optimization problems. Swarm and
  Evolutionary Computation 54:100647

\bibitem[{Zhao et~al(2012)Zhao, Suganthan, and Zhang}]{ENSMOEAD}
Zhao SZ, Suganthan PN, Zhang Q (2012) Decomposition-based multiobjective
  evolutionary algorithm with an ensemble of neighborhood sizes. IEEE
  Transactions on Evolutionary Computation 16(3):442--446

\bibitem[{Zhou et~al(2011)Zhou, Qu, Li, Zhao, Suganthan, and
  Zhang}]{zhou2011multiobjective}
Zhou A, Qu BY, Li H, et~al (2011) Multiobjective evolutionary algorithms: A
  survey of the state of the art. Swarm and evolutionary computation
  1(1):32--49

\bibitem[{Zitzler and K{\"{u}}nzli(2004)}]{ZitzlerK04}
Zitzler E, K{\"{u}}nzli S (2004) Indicator-based selection in multiobjective
  search. In: Proceedings of the 8th International Conference on Parallel
  Problem Solving from Nature, {PPSN}'2004, Birmingham, UK, pp 832--842

\bibitem[{Zitzler and Thiele(1998)}]{num22}
Zitzler E, Thiele L (1998) Multiobjective optimization using evolutionary
  algorithms—a comparative case study. In: International conference on
  parallel problem solving from nature, Springer, pp 292--301

\bibitem[{Zitzler et~al(2000)Zitzler, Deb, and Thiele}]{num16}
Zitzler E, Deb K, Thiele L (2000) Comparison of multiobjective evolutionary
  algorithms: Empirical results. Evolutionary computation 8(2):173--195

\bibitem[{Zitzler et~al(2001)Zitzler, Laumanns, and Thiele}]{SPEA2}
Zitzler E, Laumanns M, Thiele L (2001) Spea2: Improving the strength pareto
  evolutionary algorithm. TIK-report 103

\bibitem[{Zitzler et~al(2004)Zitzler, K{\"u}nzli et~al}]{IBEA}
Zitzler E, K{\"u}nzli S, et~al (2004) Indicator-based selection in
  multiobjective search. In: PPSN, Springer, pp 832--842

\end{thebibliography}
%% if required, the content of .bbl file can be included here once bbl is generated
%%\input sn-article.bbl

\begin{appendices}
	
	\section{Results in terms of IGD}
	
	The following tables are based on the results of comparative tests under the IGD. Tabel~\ref{IGD comparison} is the results for ensemble MOEAs and Table~\ref{IGD comparison basic} is the results for the foundation algorithms.
	% Table generated by Excel2LaTeX from sheet 'IGD'
	
	% Table generated by Excel2LaTeX from sheet '2'
	\begin{sidewaystable*}
		\centering
		\caption{The mean $\pm$ variance of the IGD values across the 30 runs over the testing set.
			For each testing problem, the best performance is indicated in bold (note for IGD, the smaller, the better).
			$\dagger$ indicates that the performance of the algorithm is not significantly different from the performance of MOEAs/PAP (according to a Wilcoxon?s bilateral rank sum test with $p=0.05$.
			The significance test of MOEAs/PAP against the ensemble MOEAs is summarized in the win-draw-loss (W-D-L) counts.}
		\scalebox{0.75}{
			\begin{tabular}{@{}lccccccc@{}}
				\toprule
				Problem& MOEAs/PAP        & EF-PD(\textbf{\textit{Nsize}})        & EF-PD(\textbf{\textit{Ngen}})        & MOEA/D-TAOS(\textbf{\textit{Nsize}})        & MOEA/D-TAOS(\textbf{\textit{Ngen}})        & NSGA-II/MOE(\textbf{\textit{Nsize}})        & NSGA-II/MOE(\textbf{\textit{Ngen}})   \\
				\midrule
				UF1   & 2.552E-02 $\pm$ 3.54E-05 & \textbf{2.132E-02}$\pm$ 4.44E-05 & 4.539E-02 $\pm$ 4.15E-03 & 4.528E-02 $\pm$ 6.40E-05 & 3.111E-02$\pm$ 7.18E-06 & 8.997E-02 $\pm$2.08E-04 & 7.008E-02 $\pm$ 5.99E-04 \\
				\midrule
				UF4   & \textbf{6.095E-03} $\pm$ 9.44E-08 & 7.326E-03 $\pm$ 3.36E-07 & 7.288E-03 $\pm$ 3.08E-07 & 7.423E-03 $\pm$ 6.49E-07 & 6.864E-03 $\pm$ 2.96E-07 & 7.123E-03 $\pm$ 6.88E-07 & 6.275E-03 $\pm$ 5.84E-08 \\
				\midrule
				UF6   & \textbf{2.760E-01} $\pm$ 7.34E-04 & 3.834E-01 $\pm$ 2.19E-02 & 5.829E-01 $\pm$ 6.42E-03 & 3.018E-01 $\pm$ 1.68E-03 & 3.516E-01 $\pm$ 6.02E-03 & 4.426E-01$\pm$ 1.52E-02 & 4.745E-01 $\pm$ 6.58E-03 \\
				\midrule
				UF7   & 2.029E-02 $\pm$ 7.72E-05 & 1.976E-02 $\pm$ 3.34E-05$\dagger$ & 2.146E-01 $\pm$ 7.10E-02 $\dagger$& \textbf{1.625E-02} $\pm$ 5.05E-06 & 1.658E-02 $\pm$ 3.58E-06 & 5.919E-02 $\pm$ 2.30E-03 & 4.937E-02 $\pm$ 4.02E-03 \\
				\midrule
				UF8   & \textbf{1.412E-01} $\pm$ 3.06E-04 & 2.128E-01 $\pm$ 1.83E-02 & 3.173E-01$\pm$ 5.78E-02 & 1.631E-01 $\pm$1.05E-03 & 1.784E-01 $\pm$5.29E-04 & 2.369E-01$\pm$ 1.53E-03 & 2.233E-01 $\pm$ 1.73E-03 \\
				\midrule
				WFG1  & 3.914E-01 $\pm$ 1.65E-03 & 3.166E-01 $\pm$ 2.15E-03 & 2.427E-01 $\pm$ 7.81E-05 & 2.417E-01 $\pm$ 2.52E-04 & 2.277E-01 $\pm$ 1.91E-04 & 2.110E-01$\pm$ 1.41E-03 & \textbf{1.848E-01} $\pm$ 1.79E-04 \\
				\midrule
				WFG5  & \textbf{2.282E-01} $\pm$ 1.59E-05 & 2.336E-01 $\pm$ 6.30E-05 & 2.383E-01 $\pm$ 8.66E-05 & 2.389E-01 $\pm$ 8.72E-05 & 2.399E-01 $\pm$ 6.13E-05 & 2.379E-01 $\pm$ 1.14E-04 & 2.283E-01 $\pm$ 2.40E-05$\dagger$ \\
				\midrule
				WFG7  & 1.194E+00 $\pm$ 5.67E-05 & 1.340E+00 $\pm$ 5.66E-04 & 1.322E+00 $\pm$ 8.30E-05 & 1.322E+00 $\pm$ 6.56E-04 & 1.294E+00 $\pm$ 1.40E-04 & 1.249E+00 $\pm$ 1.08E-03 & \textbf{1.175E+00} $\pm$ 3.28E-05 \\
				\midrule
				WFG8  & 2.185E+00 $\pm$ 3.57E-06 & 2.187E+00 $\pm$ 2.86E-05$\dagger$ & 2.180E+00 $\pm$ 2.98E-06 & 2.191E+00 $\pm$ 1.88E-04 $\dagger$& 2.184E+00 $\pm$ 3.89E-05 & 2.219E+00$\pm$ 9.00E-05 & \textbf{2.158E+00} $\pm$ 9.60E-06 \\
				\midrule
				DTLZ1 & 3.483E-02 $\pm$ 4.81E-04 & 1.817E-02 $\pm$ 4.02E-03 & 3.550E-03 $\pm$ 1.00E-07 & 3.443E-03 $\pm$ 4.88E-08 & 3.393E-03 $\pm$4.51E-07 & \textbf{3.259E-03} $\pm$ 2.75E-07 & 3.407E-03 $\pm$ 1.01E-06 \\
				\midrule
				DTLZ2 & \textbf{5.636E-03} $\pm$ 6.11E-08 & 7.778E-03 $\pm$ 5.51E-07 & 7.458E-03$\pm$ 8.40E-07 & 7.484E-03 $\pm$ 3.45E-07 & 6.948E-03 $\pm$ 1.40E-07 & 7.287E-03 $\pm$ 3.35E-07 & 5.642E-03 $\pm$ 7.32E-08$\dagger$ \\
				\midrule
				DTLZ5 & \textbf{5.669E-03}$\pm$ 6.23E-08 & 7.847E-03 $\pm$ 6.87E-07 & 7.141E-03 $\pm$ 2.56E-07 & 7.675E-03 $\pm$ 8.15E-07 & 7.037E-03 $\pm$ 3.23E-07 & 7.704E-03 $\pm$ 8.82E-07 & 5.708E-03 $\pm$ 1.05E-07 $\dagger$\\
				\midrule
				DTLZ7 & 6.641E-03 $\pm$ 3.46E-07 & 1.058E-02 $\pm$ 4.43E-06 & 5.673E-02 $\pm$ 1.87E-02 & 1.104E-02 $\pm$ 3.00E-06 & 1.157E-02 $\pm$ 2.68E-06 & 8.414E-03 $\pm$ 7.53E-07 & \textbf{6.522E-03} $\pm$ 1.99E-07 $\dagger$\\
				\midrule
				ZDT1  & 5.818E-03 $\pm$ 1.24E-07 & 7.610E-03 $\pm$ 1.07E-06 & 6.990E-03 $\pm$ 3.49E-07 & 7.449E-03 $\pm$ 9.57E-07 & 6.954E-03 $\pm$ 1.60E-07 & 6.691E-03 $\pm$ 5.74E-07 & \textbf{4.570E-03} $\pm$ 2.84E-08 \\
				\midrule
				ZDT2  & 6.061E-03 $\pm$ 1.87E-07 & 7.417E-03 $\pm$ 5.37E-07 & 7.364E-03 $\pm$ 6.92E-07 & 7.364E-03 $\pm$ 5.45E-07 & 6.560E-03 $\pm$ 2.51E-07 & 7.274E-03$\pm$ 8.86E-07 & \textbf{4.722E-03} $\pm$ 2.80E-08 \\
				\midrule
				ZDT6  & 4.669E-03 $\pm$ 1.34E-07 & 7.247E-03 $\pm$ 1.89E-06 & 7.324E-03 $\pm$ 7.75E-07 & 5.991E-03$\pm$ 3.29E-07 & 5.392E-03 $\pm$ 2.67E-07 & 5.650E-03 $\pm$ 1.87E-07 & \textbf{4.388E-03} $\pm$ 4.69E-07 $\dagger$\\
				\midrule
				MaOP1 & 2.998E-01 $\pm$ 3.41E-05 & 2.183E+00 $\pm$ 1.07E-03 & 2.158E+00 $\pm$ 3.70E-03 & 2.244E+00 $\pm$ 1.81E-05 & 2.244E+00 $\pm$ 9.52E-06 & 2.997E-01 $\pm$ 6.46E-05$\dagger$ & \textbf{2.941E-01}$\pm$ 5.53E-05 \\
				\midrule
				MaOP3 & \textbf{2.735E-01} $\pm$ 1.23E-08 & 9.492E-01 $\pm$ 7.28E-02 & 2.737E-01 $\pm$ 6.73E-09 & 2.736E-01$\pm$ 3.49E-09 & 2.737E-01 $\pm$ 1.77E-09 & 2.740E-01 $\pm$ 1.81E-06 & 2.740E-01 $\pm$ 4.11E-07 \\
				\midrule
				MaOP4 & 3.381E-01 $\pm$ 2.29E-03 & 3.748E-01 $\pm$ 6.28E-03 & 3.125E-01 $\pm$ 3.87E-03 $\dagger$& 2.885E-01 $\pm$ 6.33E-03 & \textbf{2.806E-01} $\pm$ 1.00E-02 $\dagger$& 5.867E-01 $\pm$ 3.03E-02 & 3.877E-01 $\pm$ 7.96E-12 \\
				\midrule
				MaOP6 & \textbf{8.074E-02}$\pm$ 1.44E-05 & 8.603E-01 $\pm$ 2.27E-01 & 7.493E-01 $\pm$ 2.49E-01 & 7.968E-01 $\pm$ 2.18E-01 & 4.694E-01 $\pm$ 1.57E-01 & 8.621E-02 $\pm$ 5.40E-05 & 9.335E-02 $\pm$ 1.24E-04 \\
				\midrule
				MaOP9 & 1.197E-01 $\pm$ 1.25E-05 & 1.379E-01 $\pm$ 5.95E-03 $\dagger$& \textbf{9.230E-02} $\pm$ 2.75E-05 & 1.202E-01 $\pm$ 8.09E-05 $\dagger$& 1.083E-01 $\pm$ 4.10E-05 & 1.544E-01 $\pm$ 3.73E-03 & 2.035E-01$\pm$ 9.02E-03 \\
				\midrule
				\midrule
				W-D-L  & - & 15-3-3 & 15-2-4 & 15-2-4 & 15-1-5 & 18-1-2 & 10-5-6 \\
				\bottomrule
			\end{tabular}%
		}
		\label{IGD comparison}%
	\end{sidewaystable*}%
	
	\begin{sidewaystable*}
		\centering
		\caption{The mean $\pm$ variance of the IGD values across the 30 runs over the testing set.
			For each testing problem, the best performance is indicated in bold (note for IGD, the smaller, the better).
			$\dagger$ indicates that the performance of the algorithm is not significantly different from the performance of MOEAs/PAP (according to a Wilcoxon?s bilateral rank sum test with $p=0.05$.
			The significance test of MOEAs/PAP against the foundation algorithms is summarized in the win-draw-loss (W-D-L) counts.}
		\scalebox{.75}{
			\begin{tabular}{@{}lccccccc@{}}
				\toprule
				Problem& MOEAs/PAP        & MOEA/D(\textbf{\textit{Ngen}})       & MOEA/D(\textbf{\textit{Nsize}})        & NSGA-II(\textbf{\textit{Ngen}})        & NSGA-II(\textbf{\textit{Nsize}})        & MOPSO(\textbf{\textit{Ngen}})        & MOPSO(\textbf{\textit{Nsize}})   \\
				\midrule
				UF1   & \textbf{2.552E-02} $\pm$ 3.54E-05 & 1.035E-01$\pm$ 1.27E-02 & 1.391E-01 $\pm$1.54E-02 & 9.560E-02$\pm$ 7.55E-04 & 9.500E-02 $\pm$ 2.19E-04 & 7.110E-02 $\pm$ 3.91E-05 & 7.262E-02$\pm$5.86E-05 \\
				\midrule
				UF4   & \textbf{6.095E-03} $\pm$ 9.44E-08 & 7.277E-03$\pm$ 2.98E-07 & 7.323E-03 $\pm$ 4.61E-07 & 6.627E-03 $\pm$7.79E-08 & 7.047E-03 $\pm$ 4.17E-07 & 9.246E-03$\pm$ 6.88E-07 & 8.901E-03$\pm$ 7.11E-07 \\
				\midrule
				UF6   & \textbf{2.760E-01} $\pm$ 7.34E-04 & 6.449E-01$\pm$ 1.45E-02 & 6.021E-01 $\pm$ 5.13E-03 & 4.190E-01$\pm$ 5.02E-03 & 4.263E-01$\pm$ 4.92E-03 & 3.003E-01 $\pm$ 4.80E-04 & 3.102E-01 $\pm$ 6.53E-04 \\
				\midrule
				UF7   & \textbf{2.029E-02} $\pm$ 7.72E-05 & 4.569E-01 $\pm$ 4.35E-02 & 4.403E-01 $\pm$ 5.57E-02 & 5.699E-02 $\pm$ 3.65E-03 & 6.137E-02 $\pm$ 9.17E-04 & 2.780E-02 $\pm$ 4.24E-06 & 3.192E-02$\pm$ 8.27E-06 \\
				\midrule
				UF8   & \textbf{1.412E-01} $\pm$ 3.06E-04 & 3.278E-01 $\pm$ 5.84E-02 & 1.900E-01 $\pm$ 1.16E-02 & 2.342E-01 $\pm$ 1.30E-03 & 2.562E-01 $\pm$ 7.68E-04 & 2.023E-01$\pm$ 4.55E-03 & 1.581E-01$\pm$ 2.62E-04 \\
				\midrule
				WFG1  & 3.914E-01 $\pm$ 1.65E-03 & 2.408E-01 $\pm$ 1.98E-04 & \textbf{2.371E-01} $\pm$ 2.75E-04 & 2.539E-01 $\pm$ 1.90E-03 & 4.501E-01 $\pm$ 7.44E-03 & 1.624E+00 $\pm$ 6.80E-04 & 1.610E+00 $\pm$ 5.10E-04 \\
				\midrule
				WFG5  & 2.282E-01$\pm$ 1.59E-05 & 2.435E-01$\pm$ 9.25E-05 & 2.355E-01$\pm$ 5.48E-05 & \textbf{2.242E-01}$\pm$ 3.00E-05 & 2.327E-01 $\pm$ 6.45E-05 & 2.842E-01$\pm$ 2.71E-04 & 2.417E-01 $\pm$ 9.29E-05 \\
				\midrule
				WFG7  & 1.194E+00 $\pm$ 5.67E-05 & 1.299E+00 $\pm$ 1.83E-04 & 1.327E+00 $\pm$ 1.02E-03 & \textbf{1.171E+00} $\pm$ 4.49E-05 & 1.470E+00 $\pm$ 9.60E-03 & 1.290E+00 $\pm$ 1.57E-04 & 1.339E+00$\pm$ 1.26E-02 \\
				\midrule
				WFG8  & 2.185E+00 $\pm$3.57E-06 & 2.186E+00 $\pm$ 2.32E-05 $\dagger$& 2.196E+00$\pm$ 1.19E-04 & \textbf{2.165E+00} $\pm$ 1.56E-04 & 2.573E+00$\pm$ 3.12E-04 & 3.102E+00$\pm$ 1.35E-04 & 3.046E+00$\pm$ 4.86E-04 \\
				\midrule
				DTLZ1 & \textbf{3.483E-02} $\pm$ 4.81E-04 & 2.376E-01 $\pm$ 1.52E-01 $\dagger$& 4.616E-01 $\pm$ 2.50E-01 & 8.971E+00 $\pm$1.00E+01 & 2.190E+00$\pm$ 3.24E+00 & 1.355E+01$\pm$ 2.03E+02 & 2.301E+01 $\pm$ 2.80E+02 \\
				\midrule
				DTLZ2 & \textbf{5.636E-03} $\pm$ 6.11E-08 & 7.341E-03$\pm$ 5.81E-07 & 7.868E-03 $\pm$ 4.85E-07 & 6.286E-03 $\pm$ 9.08E-08 & 7.290E-03$\pm$ 3.32E-07 & 6.920E-03$\pm$ 1.13E-07 & 8.180E-03$\pm$ 5.86E-07 \\
				\midrule
				DTLZ5 & \textbf{5.669E-03} $\pm$ 6.23E-08 & 7.341E-03 $\pm$ 5.81E-07 & 7.868E-03$\pm$ 4.85E-07 & 6.266E-03$\pm$ 1.01E-07 & 7.346E-03 $\pm$ 3.11E-07 & 6.920E-03 $\pm$ 1.13E-07 & 8.180E-03 $\pm$ 5.86E-07 \\
				\midrule
				DTLZ7 & \textbf{6.641E-03} $\pm$ 3.46E-07 & 1.088E-02$\pm$ 1.78E-06 & 9.692E-03 $\pm$ 1.50E-06 & 7.622E-03 $\pm$ 1.07E-05 & 7.939E-03 $\pm$ 3.64E-07 & 8.145E-03$\pm$ 6.00E-07 & 9.712E-03 $\pm$ 3.21E-06 \\
				\midrule
				ZDT1  & \textbf{5.818E-03}$\pm$ 1.24E-07 & 6.815E-03 $\pm$ 2.82E-07 & 7.396E-03$\pm$ 8.78E-07 & 9.637E-03 $\pm$ 4.98E-05 & 7.248E-03$\pm$ 1.67E-06 & 6.569E-03 $\pm$ 1.43E-07 & 7.843E-03$\pm$ 5.62E-07 \\
				\midrule
				ZDT2  & \textbf{6.061E-03} $\pm$ 1.87E-07 & 6.933E-03 $\pm$3.42E-07 & 7.151E-03$\pm$ 4.82E-07 & 1.243E+00$\pm$ 2.03E+00 & 7.563E-03$\pm$ 1.23E-06 & 6.643E-03 $\pm$ 2.16E-07 & 8.160E-03 $\pm$ 8.71E-07 \\
				\midrule
				ZDT6  & \textbf{4.669E-03} $\pm$ 1.34E-07 & 5.631E-03 $\pm$ 4.66E-07 & 6.042E-03 $\pm$ 5.88E-07 & 3.075E-01 $\pm$ 5.95E-01 & 5.721E-03$\pm$ 4.27E-07 & 6.044E-03 $\pm$ 3.37E-07 & 6.591E-03 $\pm$ 3.78E-07 \\
				\midrule
				MaOP1 & 2.998E-01$\pm$ 3.41E-05 & 2.246E+00 $\pm$ 6.64E-05 & 2.247E+00 $\pm$ 6.21E-05 & \textbf{2.994E-01}$\pm$ 3.97E-05 $\dagger$& 3.038E-01$\pm$ 8.42E-05 $\dagger$& 5.934E-01$\pm$ 1.28E-02 & 7.445E-01 $\pm$ 2.79E-03 \\
				\midrule
				MaOP3 & \textbf{2.735E-01}$\pm$ 1.23E-08 & 2.738E-01 $\pm$ 1.50E-09 & 2.736E-01 $\pm$ 2.50E-09 & 2.740E-01$\pm$ 9.23E-07 $\dagger$& 2.737E-01$\pm$ 2.10E-08 & 1.234E+00$\pm$ 2.02E-01 & 1.601E+00 $\pm$ 1.84E-01 \\
				\midrule
				MaOP4 & 3.381E-01$\pm$ 2.29E-03 & \textbf{3.089E-01} $\pm$ 8.25E-03 & 3.195E-01 $\pm$ 1.72E-03$\dagger$ & 3.877E-01$\pm$ 7.49E-12 & 6.101E-01 $\pm$ 2.86E-02 & 3.877E-01$\pm$ 1.22E-10 & 3.877E-01 $\pm$ 1.08E-10 \\
				\midrule
				MaOP6 & 8.074E-02$\pm$ 1.44E-05 & 1.148E+00 $\pm$ 2.68E-02 & 1.118E+00$\pm$ 4.70E-02 & 7.957E-02 $\pm$ 1.10E-05 $\dagger$& 8.315E-02 $\pm$ 4.43E-05 $\dagger$& \textbf{6.326E-02} $\pm$ 1.90E-06 & 6.446E-02$\pm$ 2.60E-06 \\
				\midrule
				MaOP9 & 1.197E-01 $\pm$ 1.25E-05 & 2.094E-01 $\pm$ 3.24E-02$\dagger$ & 1.915E-01 $\pm$ 1.61E-02$\dagger$ & 1.303E-01 $\pm$ 1.12E-05 & 1.356E-01 $\pm$ 2.63E-03 & \textbf{8.898E-02}$\pm$ 8.25E-06 & 1.071E-01$\pm$ 5.93E-06 \\
				\midrule
				\midrule
				W-D-L  & - & 16-3-2 & 18-2-1 & 14-3-4 & 19-2-0 & 19-0-2 & 19-0-2 \\
				\bottomrule
			\end{tabular}%
		}
		\label{IGD comparison basic}%
	\end{sidewaystable*}%
	
\end{appendices}

\end{document}